\documentclass[10pt,journal,a4paper,final,oneside,twocolumn]{IEEEtran}
\usepackage{amsmath,amsfonts}
\usepackage{algorithmic}
\usepackage{algorithm}
\usepackage{array}
\usepackage{subcaption}
\usepackage{textcomp}
\usepackage{stfloats}
\usepackage{url}
\usepackage{verbatim}
\usepackage{graphicx}
\usepackage{siunitx}
\usepackage{cite}
\usepackage{amsfonts}   % 基本解决方案
\usepackage{amssymb}    % 更推荐（包含 amsfonts 并扩展符号）
\usepackage{booktabs}    % 专业表格线型
\usepackage{multirow} 
\usepackage{booktabs}
\usepackage[utf8]{inputenc}
\usepackage{ulem}
\usepackage{threeparttable}  %表格添加备注

\begin{document}
	
	\title{A Camera-Cooperative ISAC Framework for Multimodal Non-Cooperative UAVs Sensing}
	
	\author{Wenfeng Wu,~\IEEEmembership{Student Member,~IEEE},
		Luping Xiang,~\IEEEmembership{Senior Member,~IEEE},
		Kun Yang,~\IEEEmembership{Fellow,~IEEE},
		
		% <-this % stops a space
		%\thanks{This paper ...
			
			% }% <-this % stops a space
		\thanks{Wenfeng Wu, Luping Xiang, and Kun Yang are with the State Key Laboratory of Novel Software Technology, Nanjing University, Nanjing, 210008, China, Institute of Intelligent Networks and Communications (NINE) and School of Intelligent Software and Engineering, Nanjing University (Suzhou Campus), Suzhou, 215163, China (e-mail: wenfengwu@smail.nju.edu.cn; luping.xiang@nju.edu.cn; kunyang@nju.edu.cn)}
		
	}
	
	\markboth{Journal of \LaTeX\ Class Files,~Vol.~14, No.~8, August~2021}%
	{Shell \MakeLowercase{\textit{et al.}}: A Sample Article Using IEEEtran.cls for IEEE Journals}
	
	
	\maketitle
	
	\begin{abstract}
		The detection of non-cooperative unmanned aerial vehicles (UAVs) presents significant challenges for Integrated Sensing and Communication (ISAC) systems due to the inherent limitations of single-modal perception and the competition for shared communication and sensing resources. To address these challenges, this paper proposes a novel Camera-Cooperative ISAC (CC-ISAC) framework that employs multimodal sensing to enable efficient UAV beam steering and tracking. The proposed framework employs cameras for coarse-grained airspace monitoring and utilizes ISAC for fine-grained, high-precision sensing, forming a complementary perception loop that enhances both sensing accuracy and resource efficiency. Within this framework, two key modules are developed: (1) a Vision-to-Echo Data Alignment (V2EDA) model that aligns visual and echo-domain features through cross-attention mechanisms, and (2) a Multimodal Fusion-Based Estimation (MMFE) model that integrates historical multimodal data with current observations for robust state estimation. Extensive evaluations conducted on the DeepSense6G dataset demonstrate that the proposed framework achieves an average reduction of 71\% in beam steering overhead and 1.69–11.15\% in tracking overhead while maintaining high angular estimation accuracy. The CC-ISAC framework effectively mitigates resource contention between sensing and communication, enabling reliable UAV surveillance while freeing substantial system resources for additional communication tasks, thereby representing a practical advancement in ISAC system design.
		
	\end{abstract}
	
	
	\section{Introduction}
	\IEEEPARstart{T}{he} low-altitude airspace, a rapidly developing integrated economic ecosystem relying on advanced aerial platforms such as unmanned aerial vehicles (UAVs) and electric vertical take-off and landing vehicles, has garnered significant global attention~\cite{huang2024low}. Within this airspace, cooperative UAVs are typically managed by a base station (BS) and transmit real-time location data through cellular networks~\cite{javaid2023communication}. Conversely, non-cooperative UAVs present considerable threats to public safety, necessitating timely detection and tracking for effective countermeasures~\cite{10906066,10977743}.
	
	Integrated Sensing and Communication (ISAC) systems have emerged as a promising solution for establishing wireless communication networks that provide wide coverage and real-time sensing capabilities. In ISAC systems, sensing and communication (S\&C) functionalities share finite physical resources while achieving high sensing performance and substantial data throughput~\cite{9737357}. By allocating orthogonal or non-overlapping resources in the time, frequency, and spatial domains, S\&C functions can operate concurrently without interference. As a result, the ISAC BS plays a pivotal role in both communication and sensing tasks, including the detection and tracking of UAVs to mitigate unauthorized intrusions or collisions~\cite{10906066}. This dual functionality presents significant challenges in terms of resource allocation between S\&C.
	
	In ISAC systems, motion parameters, such as range, angle, and velocity, of moving objects are estimated through the transmission of ISAC signals and subsequent analysis of the reflected echoes, utilizing techniques such as time-delay measurement, beamforming, and Doppler shift analysis~\cite{khan2022detection}. However, the inherently low detect ability of non-cooperative UAVs, stemming from their typical low-altitude, low-speed, and small radar cross-section (RCS) characteristics, introduces notable challenges for ISAC systems~\cite{song2025overview}. Furthermore, the convergence of ubiquitous wireless sensing and advanced visual sensing technologies has made rapid progress, fueled by the diverse, heterogeneous information provided by various sensors in intelligent systems~\cite{cheng2023intelligent, cai2022ubiquitous, peng2025integrated}. This convergence presents a new paradigm for non-cooperative UAV sensing.
	
	\subsection{Related work}
	\subsubsection{Vision-Based UAV Sensing}
	
	Recent advancements in deep learning have significantly improved performance in small-object detection and tracking tasks~\cite{Gold-YOLO2023wang, ASF-YOLO2024kang}. Among these, the YOLO series has emerged as a prominent solution, offering an optimal balance between real-time processing and robust detection across multiple object scales~\cite{wang2024low, realtime2025liu}. For instance, in~\cite{cai2022lightweight}, a mean Average Precision (mAP) of 93.52\% was achieved at 82 frames per second (fps) on a custom UAV dataset under simple scenarios, while~\cite{guo2024global} reported 92\% accuracy at 23.6 fps using the ARD-MAV dataset under more complex environmental conditions. These results illustrate the trade-off between detection accuracy and real-time processing capabilities in vision-based UAV sensing. However, vision-based approaches remain sensitive to challenges such as illumination variation, occlusion, and background clutter.

	\subsubsection{UAV Sensing in ISAC}
	
	To mitigate the limitations posed by weak echo signals from small targets, digital beamforming has been employed. This technique dynamically adjusts the phase and amplitude of antenna elements to steer beams, enhancing the Signal-to-Noise Ratio (SNR) and spatial resolution, thus improving the reliability of UAV detection~\cite{rajab2021multi, urzaiz2020digital, haifawi2023drone}. In practical implementations, codebook-based beam sweeping strategies are commonly used~\cite{giordani2016initial}. While exhaustive sweeping is conceptually simple, it incurs substantial overhead. Alternatively, hierarchical search methods progressively refine beamwidths and are more widely adopted in practice. By employing hierarchical codebook designs~\cite{xiao2016hierarchical}, binary-tree traversal techniques can efficiently sweep the codebook layer by layer, significantly reducing the search complexity compared to exhaustive approaches.
	
	\subsubsection{Multimodal Fusion-Based Sensing}
	
	Despite significant progress, research on multimodal fusion for non-cooperative UAV detection is still in its early stages. Sensor fusion has been widely recognized as an effective method to enhance detection accuracy by compensating for the limitations of individual sensing modalities~\cite{samaras2019deep}. As a result, various studies have explored the use of multiple sensors for joint decision-making or the construction of hierarchical frameworks to improve UAV detection~\cite{svanstrom2021real, khan2022detection}.
	
	Among different fusion paradigms, radar-vision fusion has proven to be particularly promising, enabling seamless integration of wireless sensing with computer vision. By leveraging complementary multi-dimensional features while reducing redundancy, radar-vision fusion improves environmental adaptability, reliability, and fault tolerance~\cite{10138035, yao2023radar, 11165352, peng2025large}. For example, multimodal fusion at the data-level~\cite{yadav2020radar+}, decision-level~\cite{kim2023crn}, and feature-level~\cite{jiang2022t} has demonstrated superior performance over single-modal approaches.
	
	Moreover, machine learning techniques have gained traction in multimodal beam alignment and prediction tasks~\cite{xu2022computer, imran2024environment}. These methods leverage auxiliary data to improve wireless environmental awareness and enhance prediction accuracy. For instance, camera-assisted beam tracking at the BS facilitates accurate beam prediction without additional training~\cite{10476948}, while the fusion of image and GPS data enables precise estimation of UAV position and orientation~\cite{10294031}. These multimodal approaches underscore the potential of vision-assisted ISAC systems for more robust UAV sensing.
	
		\begin{table*}[htbp]
		\centering
		\begin{threeparttable}
			
			\caption{{Comparison of CC-ISAC with Prior Vision-Assisted ISAC Frameworks}}
			\label{tablecomp}
			\begin{tabular}{cccccc}  % 列数从7列改为6列
				\toprule
				Literature
				& Scenario 
				& Vision Usage 
				& Fusion Strategy 
				& Task Partitioning \textsuperscript{1}
				& Perception Offloading\textsuperscript{2} \\
				\midrule
				\multirow{2}{*}{\cite{xu2022computer}} 
				& \multirow{2}{*}{V2I} 
				& Beam selection
				& - 
				& \multirow{2}{*}{Two} 
				& \multirow{2}{*}{No} \\
				& & Beam tracking & - & & \\
				\midrule
				\multirow{2}{*}{\cite{imran2024environment}}
				& \multirow{2}{*}{V2X}
				& Transmitter identification
				& Feature-level (with Power Data)
				& \multirow{2}{*}{One} 
				& \multirow{2}{*}{No}  \\
				& & Transmitter tracking & - & & \\
				\midrule
				\multirow{2}{*}{\cite{10476948}}
				& \multirow{2}{*}{V2V } 
				& Transmitter identification
				& Feature-level (with Power Data)
				& \multirow{2}{*}{One} 
				& \multirow{2}{*}{No}  \\
				& & Transmitter tracking & - & & \\
				\midrule
				\cite{10294031} 
				& UAV
				& Beam selection
				& Feature-level (with GPS)
				& One 
				& No \\
				\midrule
				\cite{11161281} 
				& V2I
				& Beam tracking
				& Feature-level (with GPS)
				& One 
				& No \\
				\midrule
				\multirow{3}{*}{{This work}} 
				& \multirow{3}{*}{{UAV }} 
				& {Airspace detection} 
				& {-} 
				& \multirow{3}{*}{{Three}} 
				& \multirow{3}{*}{{Yes}} \\
				& & {Beam selection} 
				& {-} & & \\
				& & {Beam tracking} 
				& {Decision-level (with echo signal)} & & \\
				\bottomrule
			\end{tabular}
			\begin{tablenotes}[flushleft]  % flushleft 左对齐，默认居中
				\footnotesize  % 字号缩小，符合IEEE规范
				\item[1] Task Partitioning represents an explicit hierarchical decomposition where different modalities are assigned distinct and complementary roles (e.g., airspace detection, guided beam selection, and beam tracking), as adopted in CC-ISAC.
				\item[2] Perception offloading indicates whether coarse-grained environmental awareness is  offloaded from the ISAC sensing module to an external modality to reduce sensing overhead.
			\end{tablenotes}
		\end{threeparttable}
	\end{table*}

	\subsection{Motivation}
	Building upon the aforementioned discussions, the following conclusions are drawn:
	\begin{enumerate}
		\item \textit{Resource Allocation in Low-Altitude ISAC Deployments:} In practical ISAC systems, where aerial and terrestrial services coexist, limited spectral and spatial resources lead to inevitable contention between aerial sensing and terrestrial communication or sensing tasks. As such, efficient resource management is crucial to ensure reliable performance in both detection and communication.
		
		\item \textit{Heterogeneous Sensing Capabilities of Multiple Modalities:} Vision-based sensing provides high-resolution spatial and semantic information but is vulnerable to line-of-sight (LoS), illumination variations, and adverse weather. Conversely, ISAC-based sensing offers long-range, all-weather robustness, though it entails significant beam management overhead and lacks fine-grained visual details. Thus, multimodal fusion is essential to capitalize on the complementary strengths of each modality, enhancing detection accuracy, reliability, and overall efficiency in non-cooperative UAV sensing.
	\end{enumerate}
	
	Overall, multimodal sensing techniques offer the potential to dynamically reallocate redundant beam or antenna resources to terrestrial services or other aerial tasks, all while ensuring robust surveillance and timely detection of unauthorized UAVs. This strategy not only ensures critical perception performance but also improves overall network efficiency.
	
	\subsection{Contributions}
	
	Motivated by the preceding considerations, this work introduces a Multimodal Fusion-Based Camera-Cooperative ISAC (CC-ISAC) framework for efficient beam steering and tracking of non-cooperative UAVs. {Table~\ref{tablecomp} categorizes the key characteristics in comparison with
	those of existing studies. Existing vision-assisted ISAC approaches that primarily use visual information to improve beam alignment accuracy within a fixed sensing pipeline. The main novelty of proposed CC-ISAC lies in cross-modal task reallocation. Specifically, it adopts explicit hierarchical task partitioning to assign differentiated and complementary roles to different modalities for airspace detection, guided beam selection, and beam tracking. Meanwhile, it realizes perception offloading by transferring coarse-grained environmental awareness from the ISAC module to other modalities, thereby reducing overall sensing overhead.} To the best of our knowledge, this is the first system-level design that leverages multimodal cooperation to directly optimize ISAC resource efficiency rather than solely improving perception accuracy. The key contributions of this work are summarized as follows:
	
	\begin{enumerate}
		\item \textit{Camera-Cooperative ISAC Framework:} { A novel framework-level innovation is introduced, redefining the functional roles of heterogeneous sensors.
		CC-ISAC assigns vision as the primary sensing modality for initial airspace awareness and coarse localization, while reserving ISAC resources for collaborative fine-grained, high-precision sensing. {Specifically, vision is responsible for airspace detection; its coarse localization guides ISAC beam selection and further serves as a complementary modality to enhance beam tracking.}
		By exploiting the complementary strengths of heterogeneous modalities, the framework adopts task reallocation, converting the trade-off between sensing accuracy and ISAC time/beam resources into a trade-off between sensing accuracy and multimodal sensing capabilities.  }

		\item \textit{Enabling Multimodal Learning Modules:}  {As supporting components customized for the proposed CC-ISAC framework, we develop two learning-based modules. 	}
		We develop the Vision-to-Echo Data Alignment (V2EDA) model that learns the nonlinear mapping from image-space observations to beamspace angles. This module provides reliable coarse localization that narrows the subsequent beam search space and enables practical cooperation between spatially separated vision and ISAC sensors. 
		We further design the MMFE model, which enhances target motion state estimation through cross-modal decision fusion, thereby effectively reducing beam tracking overhead. {The proposed estimator achieves highly reliable and robust refined angle localization,  improving the long-term throughput and continuous sensing stability of the system.
		These modules collectively ensure robust continuous sensing while maximizing the resource-saving gains of the overall framework.}

		%\item \textit{Vision-to-Echo Data Alignment (V2EDA):}  {To enable practical cooperation between spatially separated vision and ISAC sensors, we develop the V2EDA model that learns the nonlinear mapping from image-space observations to beamspace angles. A dual-stream architecture with cross-attention is introduced to jointly exploit geometric bounding-box cues and semantic visual features, allowing accurate angular inference without explicit depth measurements. This module provides reliable coarse localization that narrows the subsequent beam search space and enables efficient ISAC operation.}

		%\item \textit{Multimodal Fusion-Based Estimation (MMFE):} {We further design a MMFE model that integrates historical echo and current visual observations through decision fusion and temporal encoding, reducing beam tracking overhead. By jointly exploiting temporal continuity and cross-modal complementarity, the proposed estimator improves angle prediction accuracy and robustness under channel degradation and modality failures, thereby enhancing long-term sensing stability. }
		
		\item \textit{Numerical Simulation:} Extensive simulations were conducted using the real-world DeepSense6G dataset~\cite{DeepSense}. The results demonstrate that the proposed approach effectively reduces target angle estimation RMSE and decreases beam steering time overhead by an average of 71\%. Moreover, beam tracking time overhead was reduced by 1.69\%–11.15\%, {validating the effectiveness and robustness of the proposed CC-ISAC framework under representative sensing conditions, while improving communication resource efficiency through reduced beam management overhead.
		}
		
	\end{enumerate}

	\subsection{Organization}
	
	The remainder of this paper is structured as follows. Section~\ref{sec.2} presents the system model. Section~\ref{sec.3} introduces the CC-ISAC framework. In Section~\ref{sec.4}, the V2EDA model is proposed, followed by the description of the MMFE model in Section~\ref{sec.5}. Section~\ref{sec.6} provides numerical results and analysis of the proposed algorithms. Finally, the paper concludes in Section~\ref{sec.7}.
	
	\begin{figure}
		\begin{center}
			\includegraphics[width=\linewidth]{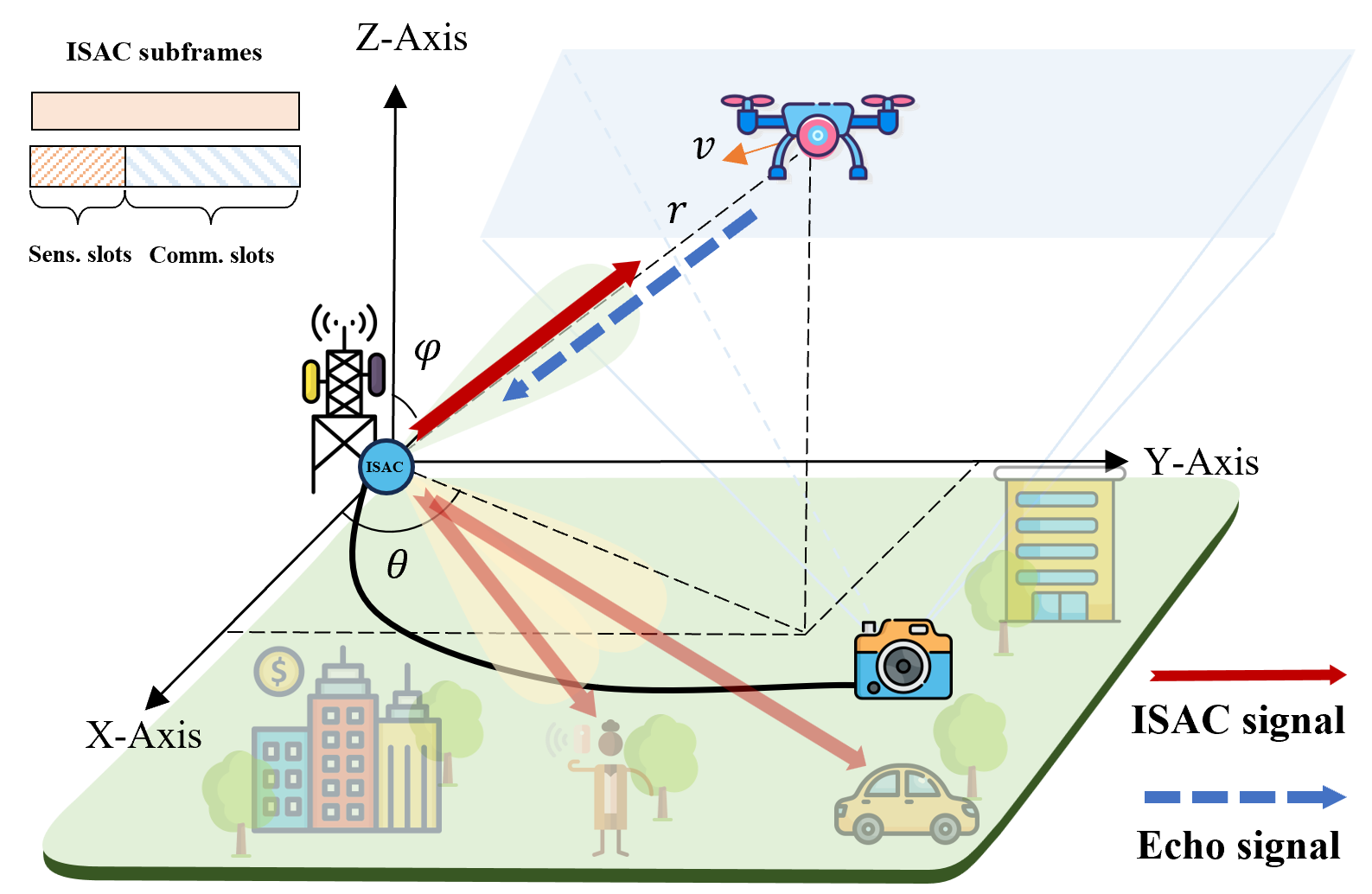}\\
			\caption{System model.}\label{fig.1}
		\end{center}
	\end{figure}
	
	\section{System model}
	\label{sec.2}
	As depicted in Fig.~\ref{fig.1}, the system model consists of an ISAC BS located at the origin on the ground. A non-cooperative UAV is assumed to be positioned at azimuth angle $\theta$, elevation angle $\phi$, and distance $d$ from the BS, and to be moving with velocity $v$. An external camera, co-located and synchronized with the BS, continuously monitors the surrounding airspace, with the UAV currently within its field of view. {Tab.~\ref{tab:isac_symbols} summarizes the main notations used throughout the system model.}
	
	The transmission frame is divided into two distinct types of time slots: sensing slots and communication slots. Enhanced efficiency in perception allows for a reduction in the number of sensing slots required for accurate sensing, thereby freeing up more slots for communication. Consequently, improved sensing efficiency directly translates into an increased overall communication capacity.
	
	\begin{table}[htbp]
		\centering
		\caption{{Main System Symbols and Descriptions}}
		\label{tab:isac_symbols}
		
		% 调整表格字体大小，根据需要可改为 \small 或 \normalsize
		\footnotesize
		\begin{tabular}{lc}
			\toprule
			\textbf{System Symbols} & \textbf{Description} \\
			\midrule
			$I$ & Image \\
			$C_I, W_I, H_I$    & Channels,width,height of the image at the camera\\
			$\mathbf{s}(t)$ & Transmitted ISAC signal \\
			$\mathbf{y}(t)$ & Received echo signal vector at the BS \\
			$p$ & Transmit power \\
			$\kappa$ & Array gain factor, $\kappa=\sqrt{N_rN_t}$ \\
			$\beta$ & Reflection coefficient of the target \\
			$\epsilon$ & RCS coefficient \\
			$d$ & Distance between the BS and UAV \\
			$v_{\parallel}$ & Radial velocity of the UAV \\
			$\mu$ & Doppler frequency shift \\
			$\tau$ & Round-trip propagation delay \\
			$f_c$ & Carrier frequency \\
			$c$ & Speed of light \\
			$\mathbf{f}$ & Transmit beamforming vector \\
			$\mathbf{z}_r(t)$ & AWGN vector \\
			$N_t$ & Number of transmit antennas \\
			$N_r$ & Number of receive antennas \\
			$\theta$ & Azimuth angle of the UAV \\
			$\phi$ & Elevation angle of the UAV \\
			$\mathbf{a}(\theta,\phi)$ & Transmit steering vector \\
			$\mathbf{b}(\theta,\phi)$ & Receive steering vector \\
			$s$    &  Codebook level\\
			$K$    &  Total number of beams in $s$ level\\
			$(m_h,m_v)$   &  Codebook index pairs \\
			\bottomrule
		\end{tabular}
	\end{table}

	\subsection{Visual Signal Model}
	
	A fixed-perspective third-party camera is utilized to provide continuous coverage of the target airspace. Let $\mathrm{I} \in \mathbb{R}^{C_I \times W_I \times H_I}$ represent the acquired image containing a single UAV target, where $C_I$, $W_I$, and $H_I$ denote the number of channels, width, and height of the image, respectively.

	\subsection{ISAC Signal Model}
	The BS incorporates a millimeter-wave massive MIMO array that supports both communication and sensing tasks concurrently. Given that UAVs typically operate at significantly higher altitudes than ground stations, it is assumed that LoS propagation is the dominant factor for the UAV-BS communication channels. The BS uses a uniform planar array (UPA) with $N_t$ transmit and $N_r$ receive antennas arranged with standard half-wavelength spacing. Operating in ISAC mode without time or frequency division, the BS transmits the ISAC signal $\mathbf{s}(t)$. When the UAV is modeled as a point target, the received echo signal at the BS is given by the following expression:
	\begin{equation}
		\label{eq.1}
		\mathbf{y}(t) = \kappa \sqrt{p}  \beta e^{j2\pi \mu t} \mathbf{b}(\theta,\phi)\mathbf{a}^H(\theta,\phi) \mathbf{f} \mathbf{s}(t - \tau) + \mathbf{z}_r(t),
	\end{equation}
	where $\mathbf{y}(t) \in \mathbb{C}^{N_r \times 1}$ represents the received signal vector, $\kappa = \sqrt{N_r N_t}$ is the array gain factor, and $p$ is the transmit power. The reflection coefficient $\beta = \epsilon (2d)^{-2}$ depends on the complex RCS $\epsilon$ and the relative distance $d$. The Doppler shift $\mu = 2v_\parallel f_c c^{-1}$ and round-trip delay $\tau = 2d c^{-1}$ are determined by the UAV's radial velocity $v_\parallel$ and its distance $d$. The beamforming vector $\mathbf{f}$ steers the transmission, and $\mathbf{z}_r(t) \sim \mathcal{CN}(0, \sigma^2 \mathbf{I})$ represents circularly symmetric complex additive white Gaussian noise (AWGN), with the transmit SNR defined as $p / \sigma^2$.
	
	Additionally, the transmit and receive steering vectors are structured as Kronecker products, as shown in the following equations: 
	\begin{align}
		\mathbf{a}(\theta,\phi) = \mathbf{v}_{N_t}(\theta,\phi) \otimes \mathbf{u}_{N_t}(\phi),\\
		\mathbf{b}(\theta,\phi) = \mathbf{v}_{N_r}(\theta,\phi) \otimes \mathbf{u}_{N_r}(\phi),
	\end{align}
	where the directional components are defined as:
	\begin{align}
		\mathbf{v}_{N_\bullet}(\theta,\phi) &= \frac{1}{\sqrt{N_{\bullet,h}}} [ 1 , e^{j\pi \sin\theta \cos\phi} , \dots , e^{j\pi(N_{\bullet,h}-1)\sin\theta \cos\phi} ], \\
		\mathbf{u}_{N_\bullet}(\phi) &= \frac{1}{\sqrt{N_{\bullet,v}}}[ 1, e^{j\pi \sin\phi} , \dots, e^{j\pi(N_{\bullet,v}-1)\sin\phi} ],
	\end{align} 
	where $N_{\bullet,h}$ and $N_{\bullet,v}$ represent the horizontal and vertical element counts in the UPA configuration.

	\subsection{Beamforming Codebook}
	The BS selects its beamforming vector $\mathbf{f}$ from a predefined multi-resolution hierarchical codebook~\cite{zhong2020novel} consisting of $s$ levels. The spatial frequencies in the azimuth and elevation domains are related to the corresponding physical angles through
	\begin{equation}
		\label{eq.6}
		\psi_h = \pi \sin(\theta) \cos(\phi), \quad \psi_v = \pi \sin(\phi),
	\end{equation}
	where $\theta$ and $\phi$ represent the azimuth and elevation angles, respectively.
	
	At level $s$, the entire angular domain is uniformly partitioned into $2^s \times 2^s$ subregions, where $(k_h, k_v)$ with $k_h, k_v \in {1, \dots, 2^{s-1}}$ denote the azimuth and elevation subregion indices, respectively. Each subregion is further divided into four finer beams, denoted by $\mathbf{f}_{(s, k_h, k_v, b_h, b_v)}$, where $b_h, b_v \in {1, 2}$. Consequently, the total number of beams in the codebook is $K = 4^s$. The corresponding beamwidths in horizontal and vertical spatial frequency domains are given by
	\begin{equation}
		\label{eq.7}
		\Delta\psi_h = \Delta\Psi_h / 2^s, \quad \Delta\psi_v = \Delta\Psi_v / 2^s.
	\end{equation}
	
	The final codebook index pairs $(m_h, m_v)$ can thus be derived as
	\begin{equation}
		m_h = 2(k_h-1)+b_h, \quad m_v = 2(k_v-1)+b_v.
	\end{equation}
	
\begin{figure*}[ht]
	\centering
	\includegraphics[width=\linewidth]{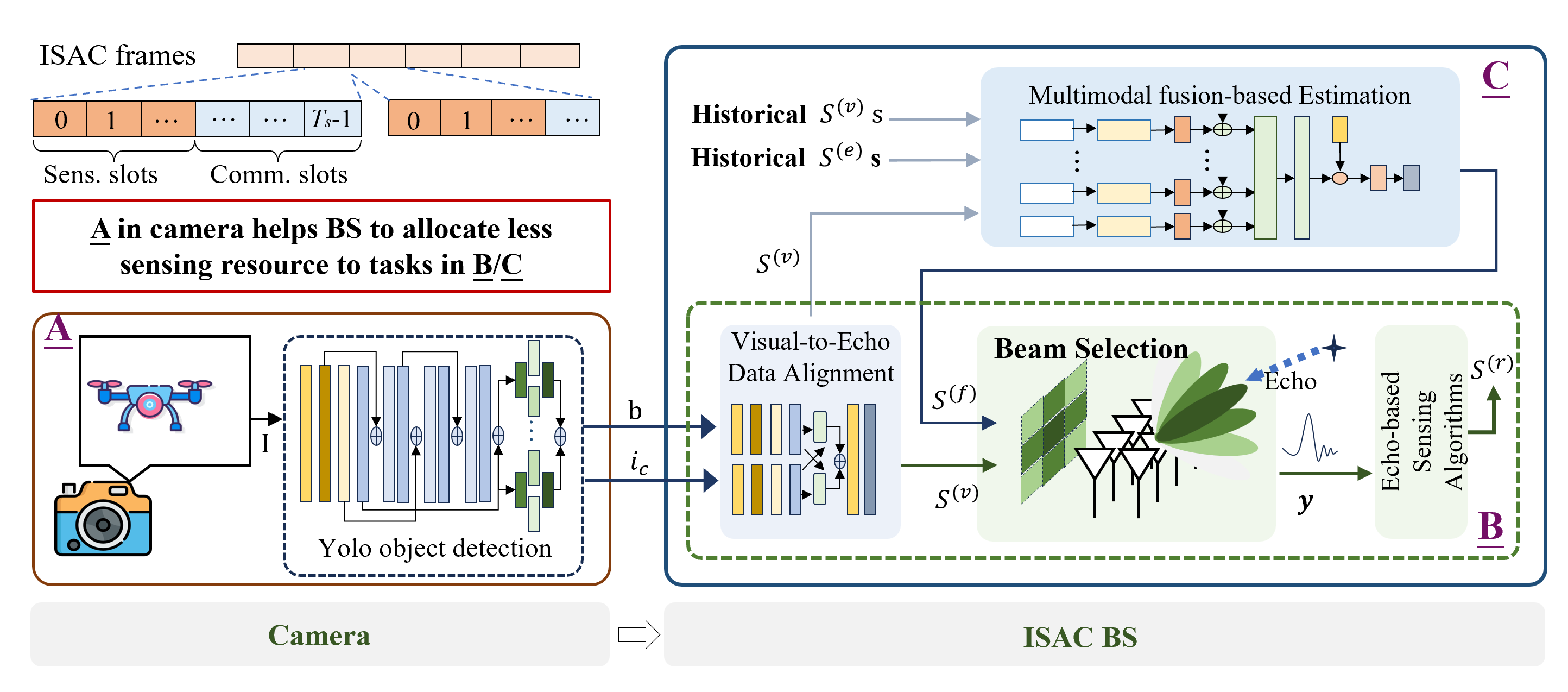}
	\caption{{CC-ISAC framework.}}
	\label{fig.2}
\end{figure*}

	\section{Camera-Cooperative ISAC Sensing Framework}
	\label{sec.3}

	This section introduces the proposed CC-ISAC framework for efficient sensing of non-cooperative UAVs, as illustrated in Fig.~\ref{fig.2}. 
	We describe the functional roles of each task and the implementation details of the sensing and perception pipeline. The design details of the two neural network models, V2EDA and MMFE, are further elaborated in Sections~\ref{sec.4} and~\ref{sec.5}, respectively. In addition, the coherence time of the ISAC link is analyzed to verify the feasibility of camera-coordinated ISAC, and the S\&C trade-off is discussed to motivate the cooperative design.
	
	The framework decomposes the overall operation into three synergistic tasks jointly executed by the external vision module and the ISAC BS:
	\begin{enumerate}
		\item \textbf{{Visual Real-Time Airspace Detection (Task A):}} The vision module captures aerial imagery and performs UAV detection using a vision-based algorithm. The resulting target information is transmitted to the ISAC BS to enable rapid situational awareness. %采用成熟的视觉检测算法对目标区域进行监控，若发现目标，即刻通过有线电缆发送关键信息给ISAC基站。
		\item \textbf{Vision-Assisted Beam Selection (Task B):} The ISAC BS processes the visual input, selects appropriate beams from the predefined codebook, performs fine-grained echo processing, and aligns the beams to establish accurate sensing links with the UAV. % 根据视觉提供的信息进行波束选择与对准，采用成熟的回波感知算法对回波进行信号处理。
		\item \textbf{Multimodal Fusion-Based Beam Tracking (Task C):} Historical sensing data and current vision observations are fused via multimodal algorithms to guide codebook selection, enabling continuous and robust beam tracking for long-term UAV localization and perception. % 当进入波束跟踪阶段，根据历史的感知结果以及当前的视觉信息，为波束跟踪的码本选择提供参考，快速定位到目标后依旧接收回波进行持续感知。
	\end{enumerate}

	By integrating an external vision system, the ISAC BS enhances its ability to localize and track non-cooperative aerial targets. Offloading the initial detection task to the vision module (Task A) allows the BS to reduce its sensing resource consumption in Tasks B and C, thereby preserving valuable spatio-temporal resources for communication and other concurrent services. This cooperative design effectively mitigates sensing overhead while improving overall network efficiency.
	
	{The proposed CC-ISAC framework is intended for scenarios where an external camera is synchronized with the ISAC base station, supported by a low-latency feedback link. It is particularly suitable for low-altitude UAV sensing with dominant LoS propagation and predefined codebooks, where visual observations can provide reliable coarse localization to guide ISAC operation. Under severe visual impairments,  the system can revert to Echo-Only tracking to maintain basic sensing capability.}
	
	%\sout{The following subsections describe each task in further detail.}
	
	\subsection{{Visual Real-Time Airspace Detection}}
	
	The YOLO object detection algorithm is adopted for UAV detection due to its robustness and computational efficiency. This module provides detection bounding box and cropped image captures that complement prior information. {The bounding box serve as coarse spatial cues of UAV location.}
	The cropped image captures the UAV’s local region, preserving texture and structural characteristics that convey information about the target’s pose, scale, and potentially even type. These extracted visual features, along with the bounding box data, are transmitted to the ISAC BS via a low-latency feedback link.
	
	Given an input RGB image $\mathbf{I}$, the detector outputs a normalized bounding box vector:
	\begin{equation}
		\label{eq.9}
		\mathbf{b} = \mathcal{F}_{\text{YOLOv4}}(\mathbf{I}) = [x_c, y_c, w, h]^\top,
	\end{equation}
	where $(x_c, y_c)$ represent the normalized coordinates of the bounding box center, and $(w, h)$ denote the normalized width and height ratios. This bounding box provides the target’s spatial reference for the ISAC BS.
	
	To extract target appearance features, pixel-level cropping is applied as
	\begin{equation}
			\label{eq.10}
			\mathbf{i}_{\text{c}} = \mathbf{I}\left[ \lfloor {W}_I x_b \rfloor : \lceil {W}_I x_e \rceil,\ \lfloor H_I y_b \rfloor : \lceil H_I y_e \rceil \right],
	\end{equation}
	where $\ x_b =x_c-w/2, x_e=x_c+w/2,	y_b=y_c-h/2,y_e=y_c+h/2$.

	\subsection{Vision-Assisted Beam Selection}
	\label{sec.3.2}
	
	\subsubsection{{Pipeline for visual sensing information}}
	Since the camera and BS operate in distinct coordinate systems, with visual data represented in an egocentric frame and ISAC sensing in a global reference, cross-modal alignment is required. The ISAC BS must align visual cues with its sensing domain to perform beam selection, and apply echo-based sensing algorithms for fine-grained spatial detection.

	To address cross-modal alignment, a V2EDA module is proposed (detailed in Sec.~\ref{sec.4}), which maps visual data to the BS sensing domain:
	{\begin{equation}
		S^{(\text{v})} = [\hat{\theta}^{(\text{v})}, \hat{\phi}^{(\text{v})}] = f_{\text{V2EDA}}(\mathrm{i}_{\text{c}}, \mathrm{b}),
	\end{equation}}
	where $S^{(\text{v})}$ denotes the angular estimates derived from visual observations, forming the visual sensing information (VSI).
	
	\subsubsection{{Hierarchical diffusive beam scanning strategy}}
	Based on $S^{(\text{v})}$, the BS selects a candidate beam set from the codebook and performs beam scanning to capture UAV echoes efficiently. The wavenumber-space mapping from~\eqref{eq.6}–\eqref{eq.7} is used to determine the central beam index pairs:
	\begin{equation}
		m_h^{(\text{v})} = \left\lfloor \frac{\psi_h^{(\text{v})}}{\Delta\psi_h^{(\text{v})}} + 0.5 \right\rfloor, \quad 
		m_v^{(\text{v})} = \left\lfloor \frac{\psi_v^{(\text{v})}}{\Delta\psi_v^{(\text{v})}} + 0.5 \right\rfloor,
	\end{equation}
	where $\psi_h^{(\text{v})}$ and $\psi_v^{(\text{v})}$ {represent} the wavenumber-space angle, $\Delta\psi_h^{(\text{v})}$ and $\Delta\psi_v^{(\text{v})}$ {represent} the wavenumber-space resolution.

\begin{figure}
	\centering
	\includegraphics[width=0.95\linewidth]{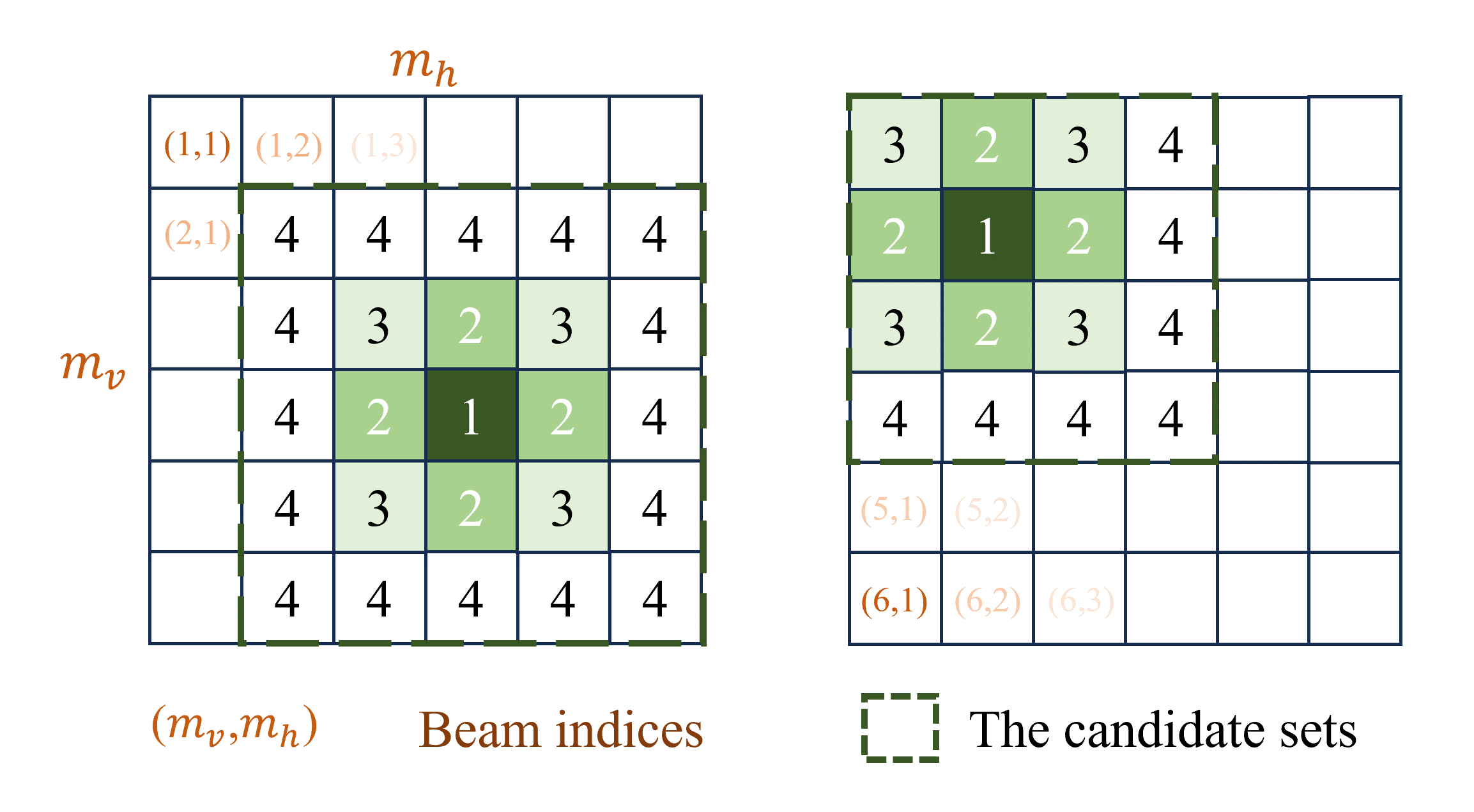}
	\caption{Diffusive beam scanning strategy.}
	\label{fig.3}
	\end{figure}

	We propose a hierarchical diffusive beam scanning strategy, in which the candidate sets $\mathcal{I}^{(1)} \rightarrow \mathcal{I}^{(2)} \rightarrow \mathcal{I}^{(3)} \rightarrow \mathcal{I}^{(4)}$ are sequentially scanned, as shown in Fig.~\ref{fig.3}.
	The candidate sets are defined as:
	\begin{equation}
		\label{eq.13}
		\begin{aligned}
			\mathcal{I}^{(1)} &= \{(m_h^{(\text{v})}, m_v^{(\text{v})})\}, \\
			\mathcal{I}^{(2)} &= \{(m_h^{(\text{v})} \pm 1, m_v^{(\text{v})}), (m_h^{(\text{v})}, m_v^{(\text{v})} \pm 1)\}, \\
			\mathcal{I}^{(3)} &= \{(m_h^{(\text{v})} \pm 1, m_v^{(\text{v})} \pm 1)\}, \\
			\mathcal{I}^{(4)} &= \{(m_h^{(\text{v})} \pm 2, m_v^{(\text{v})} \pm u),\\
			&\quad \quad (m_h^{(\text{v})} \pm u, m_v^{(\text{v})} \pm 2) \mid u \in {0,1,2}\}.
		\end{aligned}
	\end{equation}
	
	{  The ranges of $(m_h^{(\text{v})}, m_v^{(\text{v})})$ 	
		are $[1,2^s]$. 	Within the diffusion candidate set, all beam index pairs that exceed the boundary should be removed.  If the UAV remains undetected after scanning all candidate sets, this suggests either a substantial localization error or a vision detection failure. The BS then switches to an adaptive hierarchical scanning strategy to reinitialize the search~\cite{zhong2020novel}. The average number of scans for this strategy is $\log K$ . Specifically, wide beams are first used for coarse scanning, after which the beamwidth is progressively narrowed layer by layer until the optimal beam is identified.}

	As the codebook level $s$ increases, the beamwidth narrows, increasing both the angular resolution and the likelihood of misalignment from coarse localization, thereby amplifying search overhead. The average beam-search overhead, defined as the expected number of beam illuminations required until target detection, is expressed as
	{ \begin{equation}
		\label{eq.over}
		\text{Overhead} = \sum_{i=1}^{4, end}  {p_i q_i},
	\end{equation}}
	where $p_i$ denotes the probability of detecting the UAV is detected in the $i$-th candidate set, and { $p_{end}$ represents the overall failure probability, in which case the hierarchical scanning mode is triggered. Assuming that the UAV is uniformly distributed within each $i$-th candidate set, $q_i$	denotes the average number of beam scans required when detection occurs in the $i$-th candidate set. Specifically, when $i=1$, we have 
	$q=1$; for $ i \geq 2$, $q_i$ equals the cumulative size of the actual candidate sets from the first to the $i-1$-th set plus half of the size of the $i$-th candidate set. $q_{\text{end}}$	equals the total size of all candidate sets plus $\log K$. Note that when the predicted beam index lies near the boundary of the codebook, some beam pairs in the diffusive candidate sets may exceed the valid index range. These overflowed beam indices are removed before scanning, and the actual sizes of the candidate sets are used when computing $q_i$. Therefore, the beam-search overhead in Eq.~(\ref{eq.over}) reflects the effective number of scanned beams after boundary truncation.}
	
	\subsubsection{{Pipeline for echo sensing information}}

		Assuming the beam is accurately steered along the estimated angular path. { The UAV’s range and velocity are estimated using FFT-based Range–Doppler processing with spectral peak detection. 		
		Specifically, to extract the range information, the received signal is first mixed with the conjugate of the transmitted signal, yielding the intermediate-frequency (IF) signal $\mathbf{r}(t)=\mathbf{y}(t)\mathbf{s}^*(t)$. The IF signal has a single frequency tone $f_{IF}=K\tau$ and $K$ is the slope of the chirp. By applying a fast-time Fourier transform (Range FFT) to the sampled IF signal, the frequency $f_{IF}$ can be estimated, and the corresponding target distance is obtained as $d=c f_{IF}/(2K)$. 
		The range bin corresponding to the maximum spectral magnitude is selected as the estimated target range $\hat{d}^{\text(e)}$. After range processing, the echo signal samples corresponding to the detected range bin are collected across multiple pulses to form a slow-time sequence. A slow-time Fourier transform (Doppler FFT) is then applied to estimate the Doppler frequency $f_{D}$. The radial velocity can be obtained as $v=c f_{D}/{2f_c}$. The Doppler bin with the maximum spectral energy provides the velocity estimate $\hat{v}^{(\text{e})}$.

		The azimuth and elevation angles are obtained using the MUSIC algorithm~\cite{braun2011single}. Specifically, the spatial covariance matrix of the received array signals is computed and decomposed into signal and noise subspaces. The target azimuth $ \hat{\theta}^{(\text{e})},$ and elevation $\hat{\phi}^{(\text{e})}$ angles  are obtained by searching for the peak of the MUSIC spatial spectrum.}

		The ISAC echo sensing information (ESI) is then defined as
	\begin{equation}
		S^{\mathrm{(\text{e})}} = \left[ \hat{\theta}^{(\text{e})}, \hat{\phi}^{(\text{e})}, \hat{v}^{(\text{e})}, \hat{d}^{(\text{e})},... \right],
	\end{equation}
	where $S^{(\mathrm{e})}$ encapsulates the estimated motion parameters obtained from echo-domain processing.

	\subsection{Multimodal fusion-based beam tracking}
	\label{sec.3.3}
	
	% 当ISAC基站转换至跟踪状态时，其波束需要迅速转移至目标所在方向，更新跟踪目标的位置。
	
	% 首先基于视觉Net获取当前的视觉感知信息，然后，本文考虑基于多模态融合实现无人机的状态估计，提出FusionTrackNet。基于历史视觉感知信息和回波感知信息，通过融合两种模态在时间维度上的信息，预测当前状态，并通过当前的视觉感知信息进行修正。将估计的融合感知信息S^(f) 用于波束选择。
	
	% 尽管在跟踪部分多采用了深度状态估计这一过程，但是有了多模态融合信息的辅助下，提供基站更准确的无人机状态估计，能够进一步降低ISAC用于感知的波束开销。
	
	% FusionTrackNet的具体设计在第4节介绍。  通常会根据历史的感知数据进行波束预测，减少初始波束对准的开销。
	Once the ISAC BS transitions into the tracking phase, its beam must be dynamically and rapidly steered toward the target to continuously update the UAV’s position. Conventional tracking approaches typically rely on historical sensing data to reduce the overhead associated with initial beam steering; however, such methods often suffer from accumulated prediction errors over time, resulting in degraded tracking accuracy. In the proposed CC-ISAC framework, the BS estimates the UAV’s angular state for beam selection by jointly exploiting historical coarse- and fine-grained sensing data, while incorporating prior vision-derived information. Although the underlying state estimation process is inherently complex, the integration of multimodal fusion significantly enhances UAV state prediction accuracy, thereby reducing beam steering latency and improving the S\&C efficiency of the ISAC system.
	
	%\textcolor{blue}{Specifically, the current VSI $S_t^{(\text{v})}$ is first converted by the V2EDA module.  \sout{Conventional methods typically rely solely on historical sensing data for beam prediction to reduce the overhead of initial beam steering.}
		%We proposed  MMFE module (detailed in Sec.~\ref{sec.5}) to estimate UAV angle}. \sout{By integrating historical VSI and ESI across the temporal dimension, MMFE predicts the current UAV state and refines it using real-time VSI.} The multimodal-based sensing information (MSI), denoted as $S^{(f)}$, is output and then used for beam selection:
	Specifically, the current VSI, denoted as $S_t^{(\text{v})}$, is first transformed via the V2EDA module. Subsequently, the MMFE module, introduced in detail in Sec.~\ref{sec.5}, is employed to estimate the UAV’s angular parameters. The resulting multimodal sensing information (MSI), expressed as $S_t^{(\text{f})}$, is then utilized for adaptive beam selection, formulated as
	\begin{equation}
		\begin{aligned}
			S^{(\text{f})}_t &= \left[ \tilde{\theta},\tilde{\phi} \right]_t
			= f_{\text{MMFE}}(S_i^{(\text{v})}, S_j^{(\text{e})}), \\
			i &= [t-P, t-P+1, \dots, t],\\
			j &= [t-P, t-P+1, \dots, t-1],
		\end{aligned}
	\end{equation}
	where $S_i^{(\text{v})}$ and $S_j^{(\text{e})}$ denote sequences of visual and echo-domain sensing information across the most recent $P$ time slots, respectively. The detailed architecture and operation of the MMFE model are elaborated in Sec.~\ref{sec.5}.
	
	%\sout{Although the tracking process involves complex state estimation, the incorporation of multimodal fusion enables more accurate UAV state predictions at the base station, thereby further reducing the beam steering overhead in ISAC systems. }
	
	Finally, similar to Task~B, the UAV is subsequently sensed through beamforming. The echo-based sensing module outputs the current ESI $S_t^{(\text{e})}$, which is then utilized for continuous UAV tracking and state refinement.

	\subsection{Discussion on Coherence Time}
	\label{sec.3.4}
	\begin{figure}[htbp]
		\centering
		% 左子图
		\subfloat[$d=100$(m)]{
			\includegraphics[width=0.46\linewidth]{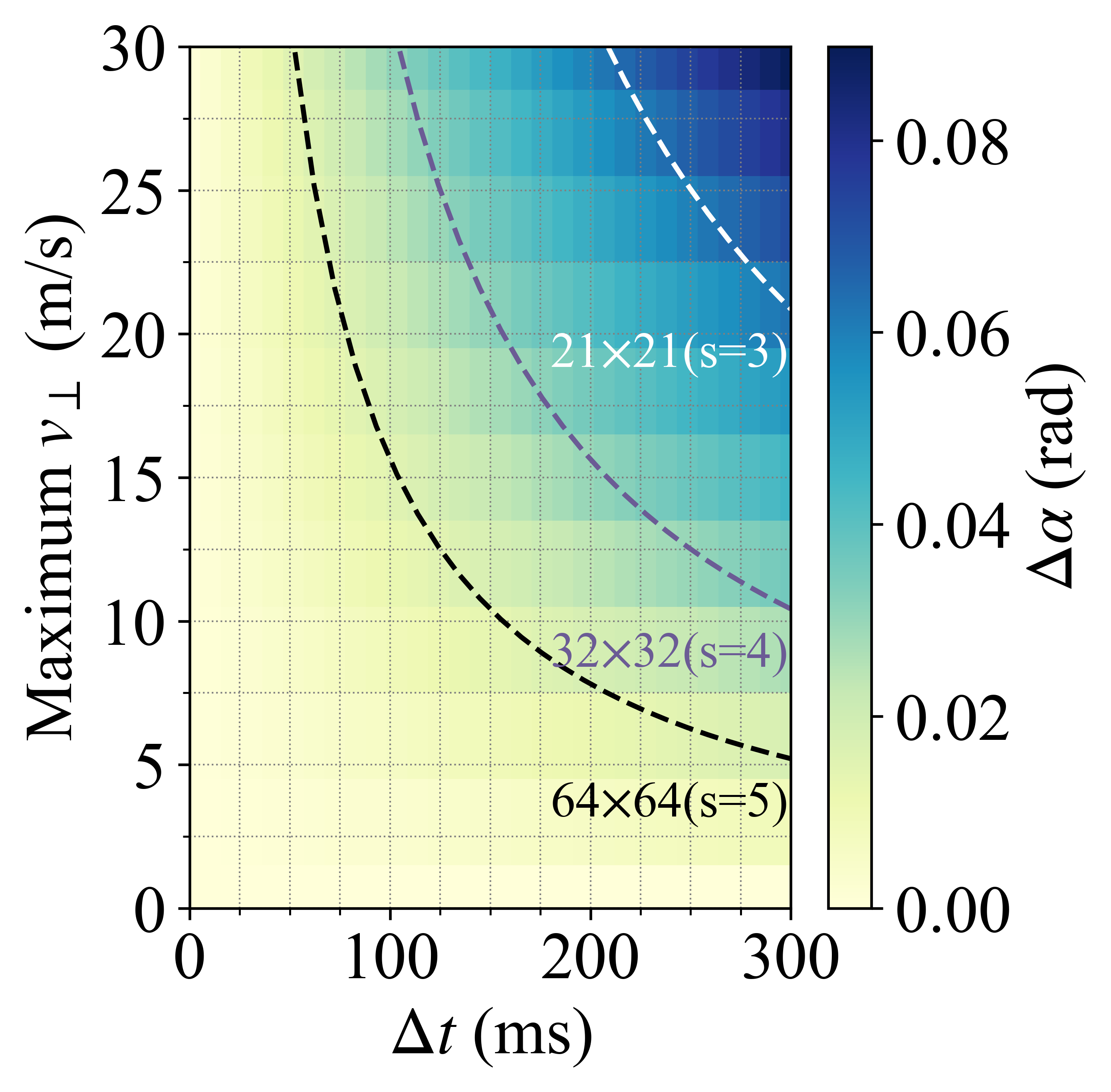}
		}
		\hfill   % 让两图之间留一点空隙
		% 右子图
		\subfloat[$d=200$(m)]{
			\includegraphics[width=0.46\linewidth]{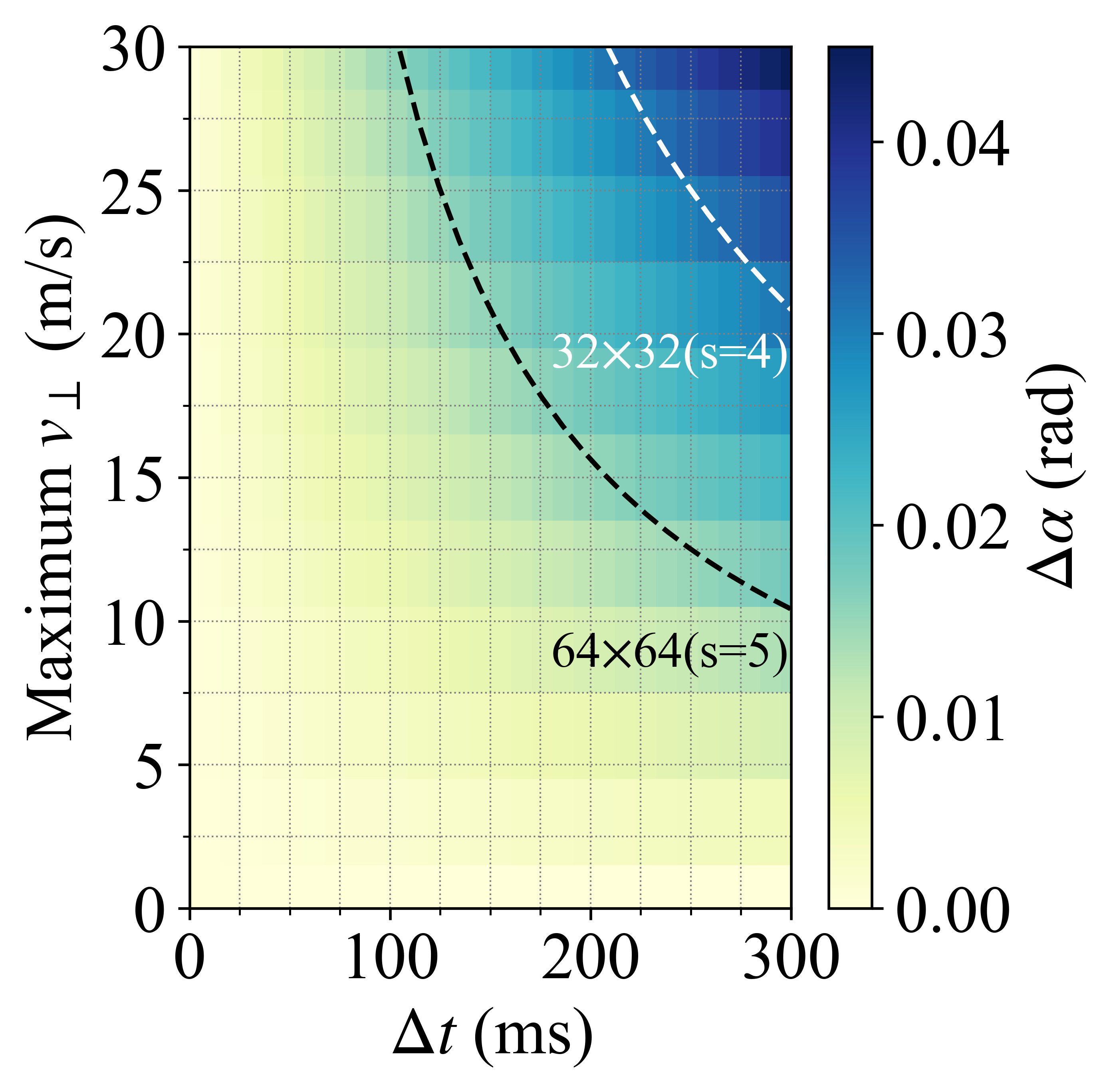}
		}
		
		\caption{Angular variation with different $\Delta t$ values for various $v_\perp$.}
		\label{fig.4}
	\end{figure}
	
	Since the activation of the directional echo-sensing beam follows the visual perception process with a latency $\Delta t$, induced by real-time monitoring, feedback transmission, and cross-modal alignment, it is necessary to analyze whether the UAV’s angular displacement during this latency could lead to beam misalignment.
	
	The predefined codebook employed in this study exhibits coupled beamspace representation across azimuth and elevation. For analytical tractability, we consider only the elevation component of angular displacement, which does not compromise the generality of the analysis.
	
	In the plane defined by the UAV’s tangential velocity $v_\perp$ and its distance $d$ from BS, the angular variation $\Delta \alpha$ over a time delay $\Delta t$ can be expressed as
	\begin{equation}
		\Delta \alpha = \arctan\left(\frac{v_\perp \Delta t}{d}\right).
	\end{equation}
	
	Mapping this angular displacement to the beamspace yields
	\begin{equation}
		\Delta \psi_\alpha = \pi \left| \sin\alpha_2 - \sin\alpha_1 \right| \leq \pi \Delta \alpha,  \quad \Delta \alpha = |\alpha_2 - \alpha_1|.
	\end{equation}
	
	Assuming an elevation beamspace coverage of $[0, \pi/2]$, the beamwidth at codebook level $s$ is given by $\Delta \psi_v^{(s)} = \pi / 2^s$. Defining the maximum tolerable angular deviation as half the beamwidth, beam misalignment can be avoided if
	\begin{equation}
		\Delta \alpha \leq 1 / 2^{(s+1)}.
	\end{equation}
	Accordingly, the corresponding coherence time is derived as
	\begin{equation}
		\Delta T_s = \frac{\tan(1 / 2^{(s+1)}) \cdot d}{v_{\perp}}.
	\end{equation}
	
	For UAV distances of $d = 100$~m and $d = 200$~m, Fig.~\ref{fig.4} illustrates the relationship between angular variation and $\Delta t$ for different tangential velocities $v_\perp$. Dashed lines in the heatmap denote the beamwidth boundaries corresponding to different codebook resolutions $s$, and the intersections between these lines and the velocity contours determine the coherence times $\Delta T_s$.
	
	Key observations from Fig.~\ref{fig.4} are summarized as follows. For a UAV at $d = 100$~m with tangential velocities below 30~m/s, the coherence time for the narrowest vertical beam ($s=5$, beamwidth $\SI{1.59}{\degree}$) is at least 50~ms, while that for a wider beam ($s=4$, beamwidth $\SI{3.17}{\degree}$) exceeds 100~ms. When the distance increases to $d = 200$~m, the coherence times approximately double: for $s=5$, $\Delta T_s \geq 100$~ms, and for $s=4$, $\Delta T_s \geq 200$~ms. In general, the coherence time increases proportionally with distance, particularly when $d > 100$~m.
	
	These results confirm the feasibility of the proposed framework, demonstrating that reliable beam steering and tracking can be achieved within realistic UAV mobility scenarios.
	
	\subsection{Discussion on the Sensing–Communication Trade-Off}
	
	In ISAC systems, S\&C inherently compete for limited spectral, spatial, and temporal resources. This trade-off can be summarized as follows: allocating more resources to sensing enhances detection and tracking performance but simultaneously reduces the resources available for communication, thereby lowering overall throughput, and vice versa.
	
	To quantify this trade-off, a simplified Time Division ISAC model~\cite{9728752} is adopted. Within a fixed frame duration $T_{\text{frame}}$, the total time resources are partitioned as
	\begin{equation}
		T_{\text{frame}} = T_{\text{Sens.}} + T_{\text{Comm.}},
	\end{equation}
	where $T_{\text{Sens.}}$ and $T_{\text{Comm.}}$ denote the durations of the S\&C slots, respectively.
	
	The total communication capacity, $C_{\text{total}}$, is proportional to the duration of the communication slot:
	\begin{equation}
		\label{eq.22}
		C_{\text{total}} \propto T_{\text{Comm.}} = T_{\text{frame}} - T_{\text{Sens.}}.
	\end{equation}
	This relationship indicates that minimizing $T_{\text{Sens.}}$ is critical to maximizing $C_{\text{total}}$. The sensing duration, in turn, depends on both the required sensing accuracy (e.g., beam steering precision) and the efficiency of the employed sensing algorithm. {The proposed CC-ISAC framework can release additional communication resources for data transmission. Moreover, the enhanced communication capability facilitates low-latency multimodal information exchange between the camera and the ISAC BS, thereby improving cross-modal coordination and enhancing the overall sensing and communication efficiency of the system.}
	
	% === III.=======================================
	% =================================================================================
	\section{Vision-to-Echo Data Alignment}
	\label{sec.4}

	 As discussed in Sec.~\ref{sec.3.2}, the V2EDA module constitutes the core component of the proposed vision-assisted beam selection framework. 
	{ Its design is motivated by the heterogeneous representation mismatch between visual observations and beamspace sensing.
		
	Traditional vision-based localization methods estimate 3D target positions by combining 2D image coordinates with auxiliary depth information. However, when explicit depth measurements are unavailable, the target depth can only be weakly inferred from cues such as bounding-box size, which often results in inaccurate spatial estimation. This limitation makes it difficult to directly map image-space observations to beamspace directions using purely geometric or single-modality-based approaches.
	
	To address this issue, V2EDA introduces a cross-modal alignment mechanism, which has been widely adopted in multimodal spatial alignment to realize cross-modal feature retrieval and alignment~\cite{11127186}. The heterogeneous nature of visual and beamspace representations fundamentally necessitates a conditional and query-dependent interaction mechanism. Specifically, the mapping from image-space features to beamspace directions is highly scene-dependent, as the same visual appearance may correspond to different spatial directions under varying camera viewpoints, target distances, and environmental conditions. Instead of assuming a fixed correspondence between visual and beamspace features, cross-attention enables the model to perform cross-modal feature retrieval, where beamspace queries dynamically retrieve relevant visual cues from image representations. Through this retrieval process, the model can align heterogeneous features across modalities and establish a flexible mapping between image-space observations and candidate beam directions.} This section details the network architecture and
	training methodology of the V2EDA model.
	
	\begin{figure}
	\centering
	\includegraphics[width=1\linewidth]{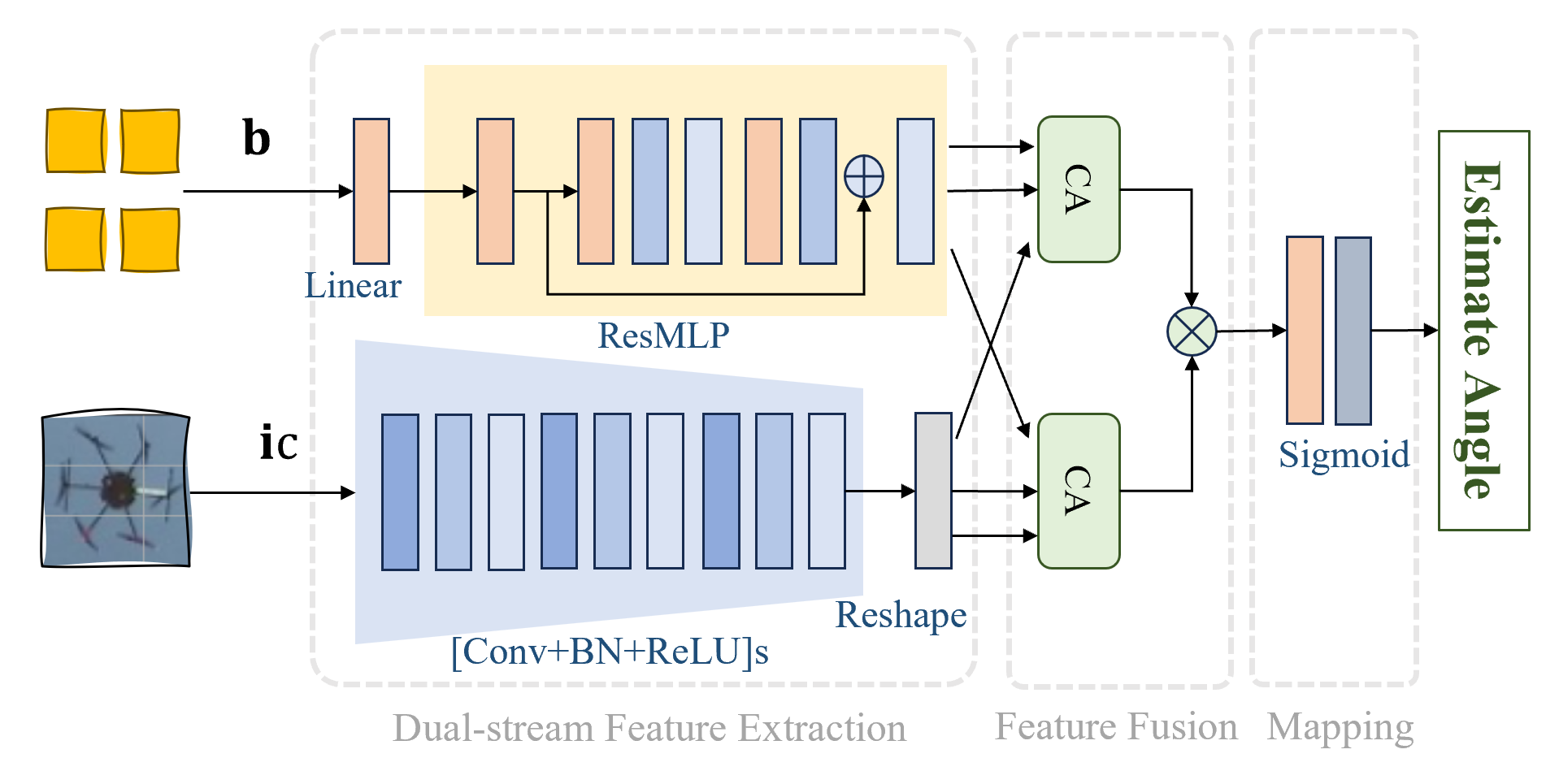}
	\caption{Architecture of the proposed V2EDA model.}
	\label{fig.5}
	\end{figure}
	
	{\begin{algorithm}[t]
		\caption{V2EDA: function $f_{\text{V2EDA}}$}
		\label{alg:mhattention1}
		
		\begin{algorithmic}[1]
			
			\item[] \textbf{Input:} 
			
			\quad \quad $\mathbf{b} \in \mathbb{R}^{4}$, UAV bounding box vector.
			
			\quad \quad $\mathbf{i}_{\text{c}} \in \mathbb{R}^{\mathbf{W}_I w \times H_I h}$, the cropped image.
			
			\item[] \textbf{Output:}
			$\hat{\theta}^{(\text{v})},\hat{\phi}^{(\text{v})} \in \mathbb{R}$, the UAV location.
			
			\item[] \textbf{Hyperparameters:}
			number of latent units $l_{\text{vis}}$, range of camera field $\theta_{\max}, \theta_{\min}, \phi_{\max}, \phi_{\min}$.
			
			\item[] \textbf{Parameters:}
			
			\quad \quad $\mathbf{W}_{b}^{\text{emb}} \in \mathbb{R}^{4 \times l_{\text{vis}}} $, the box vector embedding matrix.
			
			\quad \quad $\mathbf{W}_Q^{\text{geo}},\mathbf{W}_K^{\text{geo}},\mathbf{W}_V^{\text{geo}},\mathbf{W}_Q^{\text{sem}},\mathbf{W}_K^{\text{sem}},\mathbf{W}_V^{\text{sem}} \in \mathbb{R}^{l_{\text{vis}}^2}$, the linear projection in  $\mathit{CA}$.
			
			\quad \quad $W^{\text{pro}} \in \mathbb{R}^{l_{\text{vis}}} $, the linear projection in mapping.  
			
			\quad \quad $\mathit{ResMLP}$, $\mathit{CNN}$.
			
			\STATE  $\mathbf{f}_{\text{geo}} \leftarrow \mathit{ResMLP}(\mathbf{W}_b^{\text{emb}} \mathbf{b})$
			\STATE  $\mathbf{f}_{\text{sem}} \leftarrow \mathit{CNN}(\mathbf{i}_{\text{c}})$
			\STATE  $\mathbf{f}_{1} \leftarrow \mathit{CA}(\mathbf{f}_{\text{geo}}, \mathbf{f}_{\text{sem}} \mid \mathbf{W}_Q^{\text{geo}},\mathbf{W}_K^{\text{geo}},\mathbf{W}_V^{\text{sem}} )$~(\ref{eqcv23})
			\STATE  $\mathbf{f}_{2} \leftarrow \mathit{CA}(\mathbf{f}_{\text{geo}}, \mathbf{f}_{\text{sem}} \mid \mathbf{W}_Q^{\text{sem}},\mathbf{W}_K^{\text{sem}},\mathbf{W}_V^{\text{geo}} )$~(\ref{eqcv24})
			\STATE $\mathbf{f}_{\text{fuse}} = f_1 \cdot f_2 \in \mathbb{R}^{l_{\text{vis}}}$  ~(\ref{eqcv25})
			\STATE $a,b \leftarrow Sigmoid(W^{\text{pro}} \mathbf{f}_{\text{fuse}})$
			\RETURN 
			
			$\hat{\theta}^{(\text{v})}, \hat{\phi}^{(\text{v})} \leftarrow \text{}(a,b \mid \theta_{\max}, \theta_{\min}, \phi_{\max}, \phi_{\min}) $~(\ref{eqcv26})(\ref{eqcv27})
			
		\end{algorithmic}
	\end{algorithm}}

	\subsection{Model Architecture}
	The overall architecture of the proposed V2EDA model is illustrated in Fig.~\ref{fig.5} and Alg.~\ref{alg:mhattention1}.

	\subsubsection{Dual-stream Feature Extraction} 
	The input bounding-box vector $\mathbf{b}$ is first embedded into a high-dimensional feature space {with parameter $\mathbf{W}_{b}^{\text{emb}}$}, from which structured positional features $\mathbf{f}{\text{geo}} \in \mathbb{R}^{l{\text{vis}}}$ are extracted using a residual multilayer perceptron (ResMLP)~\cite{ma2022rethinking}. Simultaneously, the cropped image patch $\mathbf{i}_\text{c}$ is processed through a deep convolutional neural network {(CNN)} to extract environmental semantic features $\mathbf{f}{\text{sem}} \in \mathbb{R}^{l_{\text{vis}}}$, thereby capturing contextual and structural visual information.
	
	\subsubsection{Cross-Attention-Based Visual Feature Fusion}  
	A cross-attention (CA) mechanism is employed to integrate the geometric and semantic features into a unified latent representation $\mathbf{f}_{\text{fuse}}\in \mathbb{R}^{l_{\text{vis}}}$. To be specific, in the top CA module, the respective query $\mathbf{Q}_1$, key $\mathbf{K}_1$, and value $\mathbf{V}_1$ matrices are derived from the output of embedding layer as 
	$\mathbf{Q}_1 = \mathbf{f}_{\text{geo}} \mathbf{W}_Q^{\text{geo}}$, 
	$\mathbf{K}_1 = \mathbf{f}_{\text{geo}} \mathbf{W}_K^{\text{geo}}$, 
	$\mathbf{V}_1 = \mathbf{f}_{\text{sem}} \mathbf{W}_V^{\text{sem}}$. 
	Here, $\mathbf{W}_Q^{\text{geo}}$,  $\mathbf{W}_K^{\text{geo}}$, and $\mathbf{W}_V^{\text{sem}}$ are learnable parameters during training, whereas the dimensionality of both $\mathbf{Q}_1$,  $\mathbf{K}_1$, and $\mathbf{V}_1$ is $l_{\text{vis}}$. Then, we can get the attention score by computing the following scaled dot-product self-attention operation equation:
	\begin{equation}
		\label{eqcv23}
		\mathbf{f}_1 = \text{Softmax}(\frac{\mathbf{Q}_1 \cdot \mathbf{K}_1^\top}{\sqrt{l_{\text{vis}}}}) \mathbf{V}_1  \in \mathbb{R}^{l_{\text{vis}}}.
	\end{equation}
	In the bottom CA module, the attention score is 
	\begin{equation}
		\label{eqcv24}
		\mathbf{f}_2 = \text{Softmax}(\frac{\mathbf{Q}_2 \cdot \mathbf{K}_1^\top}{\sqrt{l_{\text{vis}}}}) \mathbf{V}_2  \in \mathbb{R}^{l_{\text{vis}}}.
	\end{equation}
	where the vector
	$\mathbf{Q}_2 = \mathbf{f}_{\text{sem}} \mathbf{W}_Q^{\text{sem}}$, 
	$\mathbf{K}_2 = \mathbf{f}_{\text{geo}} \mathbf{W}_K^{\text{geo}}$, 
	$\mathbf{V}_2 = \mathbf{f}_{\text{geo}} \mathbf{W}_V^{\text{geo}}$. $\mathbf{W}_Q^{\text{sem}}$,  and $\mathbf{W}_K^{\text{geo}}$, and $\mathbf{W}_V^{\text{geo}}$ are learnable parameters during training.
	
	Subsequently, the fused feature is
	\begin{equation}
		\label{eqcv25}
		\mathbf{f}_{\text{fuse}} = \mathbf{f}_1 \cdot \mathbf{f}_2 \in \mathbb{R}^{l_{\text{vis}}}.
	\end{equation}

	\subsubsection{ Mapping}
	The fused feature $\mathbf{f}_{\text{fuse}}$ is projected into the base station’s central coordinate system via a linear layer {with parameter $W^{\text{pro}}$}, with Sigmoid used as the activation function, yielding unit $a\in \mathbb{R}$ or $b\in \mathbb{R}$. Due to the limited field of view of the visual sensor, i.e., azimuth $\theta_{\text{range}} \in [\theta_{\min}, \theta_{\max}]$ and elevation $\phi_{\text{range}} \in [\phi_{\min}, \phi_{\max}]$, the two units are scaled accordingly to produce the output:
	\begin{equation}
		\label{eqcv26}
		\hat{\theta}^{(\text{v})} = a \cdot (\theta_{\max} - \theta_{\min}) + \theta_{\min} ,
	\end{equation}
	\begin{equation}
		\label{eqcv27}
		\hat{\phi}^{(\text{v})} = b \cdot (\phi_{\max} - \phi_{\min}) + \phi_{\min}.
	\end{equation}
	
	\subsection{Training}
	A supervised learning strategy is adopted. The ground truth spatial coordinates ${\theta}^{(\text{g})}, {\phi}^{(\text{g})}$ are normalized, 
	which facilitates model convergence. The mean square error (MSE) is minimized as the loss function.
	
	This enables the model to autonomously learn the complex nonlinear mapping from the visual sensor coordinate system to the BS coordinate system, thereby rapidly achieving the underlying transformation for cross-modal spatial matching. However, this mapping implicitly captures the camera’s intrinsic and extrinsic parameters, and therefore the model needs to be retrained when the camera or its viewpoint is changed. {In practice, this issue can be mitigated through data augmentation, domain adaptation, or lightweight fine-tuning using a small amount of calibration data.}

	% 1） 视觉无人机检测算法会实时检测相机视野中的恶意无人机；2） 当双目镜头中的目标无人机同时被检测到后，利用特征匹配和立体几何的方法计算出物体的在相机坐标系下的坐标。结合节点的IMU和RTK-GPS，通过坐标转换就能够获得目标无人机的GPS坐标；3） 基于轨迹的目标跟踪算法会利用该坐标进行参数的初始化，并一直跟踪目标的空间位置，从而获得一段时间内的轨迹。在中心节点中，通过轨迹相似度方法将多个感知节点估计的无人机的轨迹融合到一起，输出最终的无人机轨迹。

	\section{Multimodal fusion-based estimation}
	\label{sec.5}

	\begin{figure*}
		\centering
		\includegraphics[width=0.95\linewidth]{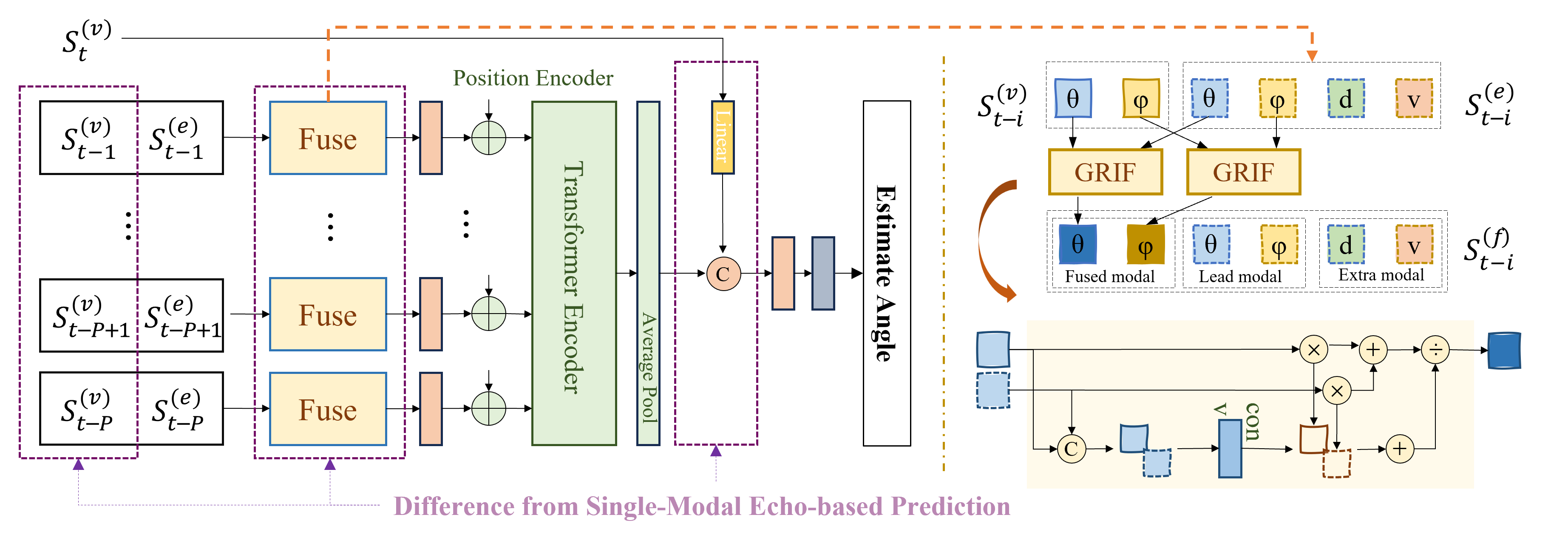}
		\caption{Architecture of the proposed MMFE model.}
		\label{fig.6}
	\end{figure*}
	
	This section details the MMFE model introduced in Sec.~\ref{sec.3.3}, including network design, training, and inference. {The MMFE module is designed to address the temporal uncertainty and modality reliability issues that arise during beam tracking in dynamic environments. Echo-based sensing provides high-resolution angular information but is susceptible to channel fading, blockage, and intermittent sensing failures. Visual observations, on the other hand, offer complementary spatial context but may suffer from occlusion, illumination changes, or detection errors. Relying on a single modality may therefore lead to unstable tracking performance.
	
	Inspired by multimodal fusion-based sensing, MMFE adopts a multimodal temporal fusion strategy to jointly exploit the temporal continuity of echo measurements and the spatial cues provided by vision. By integrating historical echo features with current visual observations, the estimator can improve prediction robustness and reduce beam tracking overhead. The resulting processing latency is small and remains} well within the feasibility bounds for UAV sensing, as established by the coherence-time analysis in Sec.~\ref{sec.3.4}. 

	{\begin{algorithm}[t]
		\caption{MMFE: function $f_{\text{MMFE}}$}
		\label{alg:mhattention3}
		
		\begin{algorithmic}[1]
			
			\item[] \textbf{Input:} 
			
			\quad \quad $\{S_{t-i}^{(\text{v})}, S_{t-i}^{(\text{e})}\}_{i=1}^P$, the historical VSI and ESI.
			
			\quad \quad $S_{t}^{(\text{v})}$, the current VSI.
			
			\item[] \textbf{Output:}
			$\tilde{\theta},\tilde{\phi} \in \mathbb{R}$, the UAV location.
			
			\item[] \textbf{Hyperparameters:}
			number of latent units $l_{\text{his}}, l_{\text{now}}$, range of camera field $\theta_{\max}, \theta_{\min}, \phi_{\max}, \phi_{\min}$.
			
			\item[] \textbf{Parameters:}
			
			\quad \quad $\mathit{Fuse}$, the fusion function in (\ref{eqfuse28}).
			
			\quad \quad $\mathit{TransEnc}$, the Transformer Encoder function.
			
			\quad \quad $\mathbf{W}_{h}^{\text{emb}} \in \mathbb{R}^{l_{\text{his}} + l_{\text{now}}} $, the VSI vector embedding matrix.
			
			\quad \quad $\mathit{MLP}$, the MLP function.
			
			\STATE  $\forall i \in \{1,\dots,P\}: S^{(f)}_{t-i} \leftarrow \mathit{Fuse}(S_{t-i}^{(\text{v})}, S_{t-i}^{(\text{e})})$
			
			\STATE  $\mathbf{f}_{\text{his}} \leftarrow \text{TransEnc}(\{S_{t-i}^{(\text{v})}, S_{t-i}^{(\text{e})}\}_{i=1}^P) \in \mathbb{R}^{l_{\text{his}}}$
			
			\STATE $\mathbf{f}_{\text{now}} \leftarrow S_t^{(\text{v})} \mathbf{W}_{h}^{\text{emb}}  \in \mathbb{R}^{l_{\text{now}}} $
			
			\STATE $a,b \leftarrow Sigmoid(\mathit{MLP}([\mathbf{f}_{\text{now}}, \mathbf{f}_{\text{his}}])) \in  \mathbb{R}$
			\RETURN 
			
			$\tilde{\theta}, \tilde{\phi} \leftarrow (a,b \mid \theta_{\max}, \theta_{\min}, \phi_{\max}, \phi_{\min}) $~(\ref{eqcv29})(\ref{eqcv30})
			
		\end{algorithmic}
	\end{algorithm}}
	
	{\begin{algorithm}[t]
		\caption{Fuse: function $f_{\text{Fuse}}$}
		\label{alg:mhattention2}
		
		\begin{algorithmic}[1]
			
			\item[] \textbf{Input:} 
			
			\quad \quad $S_{t-i}^{(\text{v})}$ consist of $\hat{\theta}^{(\text{v})}, \hat{\phi}^{(\text{v})}$.
			
			\quad \quad $S_{t}^{(\text{v})}$, consist of $\hat{\theta}^{(\text{e})}, \hat{\phi}^{(\text{e})}, \hat{v}^{(\text{e})}, \hat{d}^{(\text{e})}$.
			
			\item[] \textbf{Output:}
			$S_{t-i}^{(f)}$.
			
			\item[] \textbf{Hyperparameters:}

			\item[] \textbf{Parameters:} $\mathit{GRIF}_1, \mathit{GRIF}_2$.

			\STATE  $\hat{\theta}^{(f)} \leftarrow \mathit{GRIF}_1(\hat{\theta}^{(\text{v})}, \hat{\theta}^{(\text{e})})$
			\STATE  $\hat{\phi}^{(f)} \leftarrow \mathit{GRIF}_2(\hat{\phi}^{(\text{v})}, \hat{\phi}^{(\text{e})})$

			\RETURN 
			$S_{t-i}^{(f)} = [\hat{\theta}^{(f)}, \hat{\phi}^{(f)}, \hat{\theta}^{(\text{e})}, \hat{\phi}^{(\text{e})}, \hat{v}^{(\text{e})}, \hat{d}^{(\text{e})}] $
			
		\end{algorithmic}
	\end{algorithm}}

	\subsection{Model Design}
	As illustrated in Fig.~\ref{fig.6} and Alg.~\ref{alg:mhattention3}, MMFE consists of three components.
	\subsubsection{Multimodal Fusion}
	Given the past $P$ steps of VSI/ESI, we first form fused all $P$ samples
	\begin{equation}
		\label{eqfuse28}
		S^{(f)}_{t-i} = f_{\text{Fuse}}(S_{t-i}^{(\text{v})}, S_{t-i}^{(\text{e})}), \quad i=1, \dots, P,
	\end{equation}
	where, $f_{\text{Fuse}}$ denotes the fusion function, as shown in Alg.~{\ref{alg:mhattention2}}. Specifically, shared parameters are aligned using GRIF~\cite{9341177}, while lead modal parameters and extra useful modals such as $\hat{d}$ are concatenated. GRIF adaptively learn the fusion weights with trainable parameters. 
	
	\subsubsection{Transformer–based Temporal Encoding}
	 A Transformer encoder {(TransEnc)} captures long-range temporal dependencies over the fused sequence. { Without loss of generality, the TransEnc consists of an embedding layer, positional encoding, stacked encoder layers, and an average pooling operation. The encoder processes the fused sequence within the historical observation window and maps it to a fixed-dimensional historical representation through the encoder mapping function. After average pooling across the temporal dimension, a single feature vector of dimension ${l_{\text{his}}}$ is obtained as the historical representation $\mathbf{f}_{\text{his}}$.}

	\subsubsection{Correction Embedding and Output}
	The current VSI $S_t^{(\text{v})}$ is embedded via a linear layer, $\mathbf{f}_{\text{now}}= S_t^{(\text{v})}\mathbf{W}_{h}^{\text{emb}} \in \mathbb{R}^{l_{\text{now}}}$, where $\mathbf{W}_{h}^{\text{emb}}$ is the learnable parameter matrix. The concatenated feature $[\mathbf{f}_{\text{now}}, \mathbf{f}_{\text{his}}]$ is fed into a multi-layer perceptron (MLP) followed by a sigmoid function to produces the final angle estimate $a$ or $b$. {Similar to V2EDA model, the two units are scaled accordingly to produce the output:
	\begin{equation}
		\label{eqcv29}
		\tilde{\theta} = a \cdot (\theta_{\max} - \theta_{\min}) + \theta_{\min} ,
	\end{equation}
	\begin{equation}
		\label{eqcv30}
		\tilde{\phi} = b \cdot (\phi_{\max} - \phi_{\min}) + \phi_{\min}.
	\end{equation}}

	\subsection{Training and Inference}
	
	The output format and loss function follow the definitions in Sec.~\ref{sec.4}. During inference, the framework accounts for potential missing data scenarios:
	
	\subsubsection{Robustness to Visual Impairment}
	{For moderate visual degradations, such as partial occlusion, illumination variation, or motion blur, the robustness of V2EDA can be improved by incorporating similar impairments into the training data, which is a common strategy in vision-based learning systems. When visual observations become unreliable due to severe occlusion,  or targets moving outside the camera field of view, the framework automatically switches to a conventional single-modal echo-based prediction mode using historical ESI, rather than completely losing sensing capability, as illustrated in Fig.~\ref{fig.6}. This fallback mechanism ensures tracking continuity under severe visual impairments.}
	
	\subsubsection{Robustness to Echo Unavailability}    
	In the event of temporary echo failure, the model substitutes the missing echo input with either concurrently available visual perception data or the most recent echo measurement, thereby maintaining reliable angle estimation.

	\section{Simulation Results}
	
	\label{sec.6}
	
	In this section, we present the simulation results of the proposed V2EDA, MMFE, and CC-ISAC{, aiming to validate the rationality of the two designed modules as well as the effectiveness and feasibility of the overall CC-ISAC framework.}
	
	\subsection{Simulation Setting}
	We validate the framework on the DeepSense 6G dataset~\cite{DeepSense} curated at Arizona State University. The dataset provides synchronized RGB imagery and UAV ground-truth GPS. Data were collected outdoors with the camera oriented skyward to capture UAV trajectories. RGB frames are $960\times540$ at 30~fps, yielding clear observations with occasional, mild occlusions caused by human activity. The dataset is divided into 80\% for training and 20\% for testing, ensuring that both subsets contain samples with minor visual occlusions. The simulation parameters are summarized in Table ~\ref{tab.parameters}. The simulation parameters follow representative configurations widely used in mmWave ISAC studies unless otherwise specified. {The current evaluation should therefore be interpreted as controlled system-level validation rather than full hardware verification.}
	
	\begin{table}[htbp]
	\centering
	\caption{Simulation Parameters}
	\label{tab.parameters}
	% 关键：自动宽度、左右留白、排版美观
	\begin{tabular}{ll@{\hspace{40pt}}ll@{\hspace{20pt}}}
		\toprule
		Parameters & Value & Parameters & Value \\
		\midrule
		$s$                & 2,3,4,5,6          & $\theta_{\max}$ & $110^{\circ}$ \\
		$N_r,N_t$          & 8,8                & $\theta_{\min}$ & $150^{\circ}$\\
		$p$                & 1                  & $\phi_{\max}$   & $10^{\circ}$ \\
		$\epsilon$         & 10                 & $\phi_{\min}$   & $80^{\circ}$ \\
		$f_c$              & 28GHz              & $l_{\text{eho}}$ & 16 \\
		SNR                & 0 (default), -1, -2 & $l_{\text{his}}$ & 16 \\
		$l_{\text{vis}}$   & 16                 & $l_{\text{now}}$ & 4 \\
		$P$                & 6                  & $w$             & 0.2\\
		$t$                & 1/60fps            & $h$             & 0.2\\
		\bottomrule
	\end{tabular}
	\end{table}

	To emulate an ISAC BS situated 100~m from the camera, we synthesize the echo signal according to Eq.~(\ref{eq.1}) under a LoS channel, incorporating path loss and  AWGN with a SNR. For each UAV state over time, the azimuth, elevation, and range are estimated through Range–Doppler processing followed by the MUSIC algorithm (detailed in Sec.~\ref{sec.3.2}). Based on the geometric relationship between the camera and the ISAC BS, together with the camera field-of-view (FoV), the observable angular region of the UAV relative to the ISAC BS can be determined. Specifically, given the camera–BS distance and the FoV limits of the camera, the corresponding azimuth and elevation bounds in the ISAC coordinate system are obtained, denoted by $\theta_{\max}, \theta_{\min}, \phi_{\max}, \phi_{\min}$. The model weight parameters and normalized ratios $w,h$ are empirically selected to achieve stable training and balanced multimodal feature representation.

	\subsection{V2EDA Results}
	\label{sec.6.1}

	\begin{figure}
		
		\centering
		\begin{subfigure}{\linewidth}
			\includegraphics[width=0.98\textwidth]{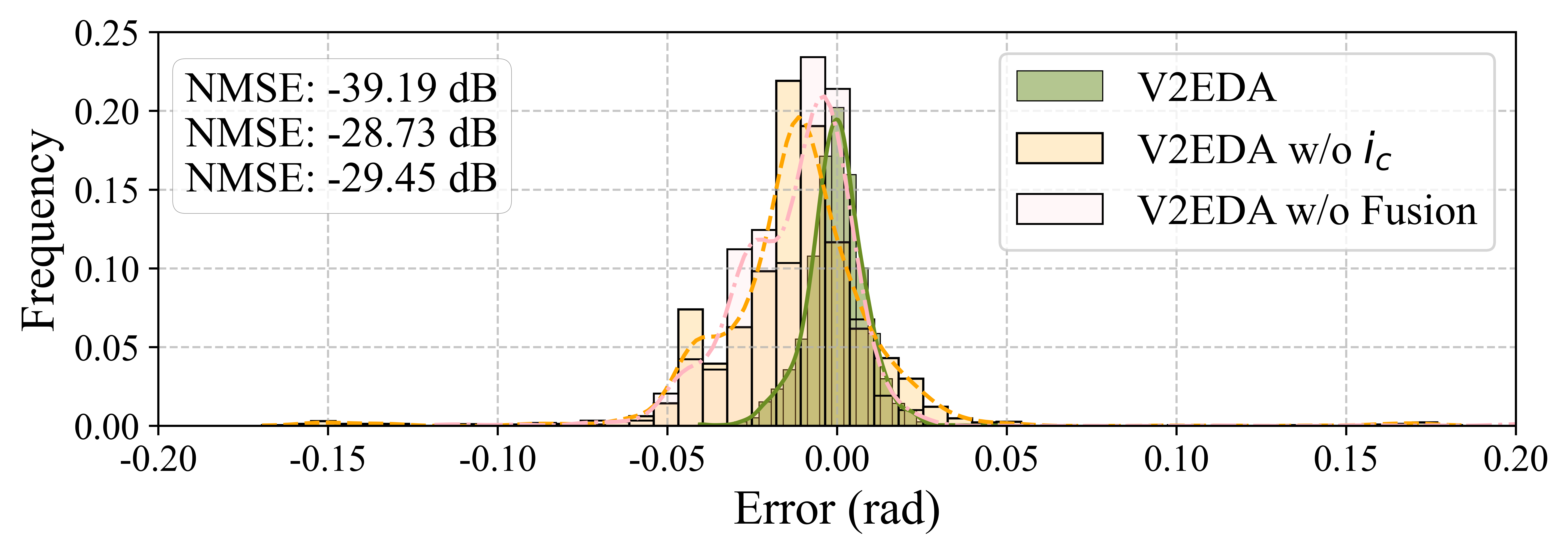}
			\caption{Azimuth}
			%\label{fig.7_1}
		\end{subfigure}
		\begin{subfigure}{\linewidth}
			\includegraphics[width=0.96\textwidth]{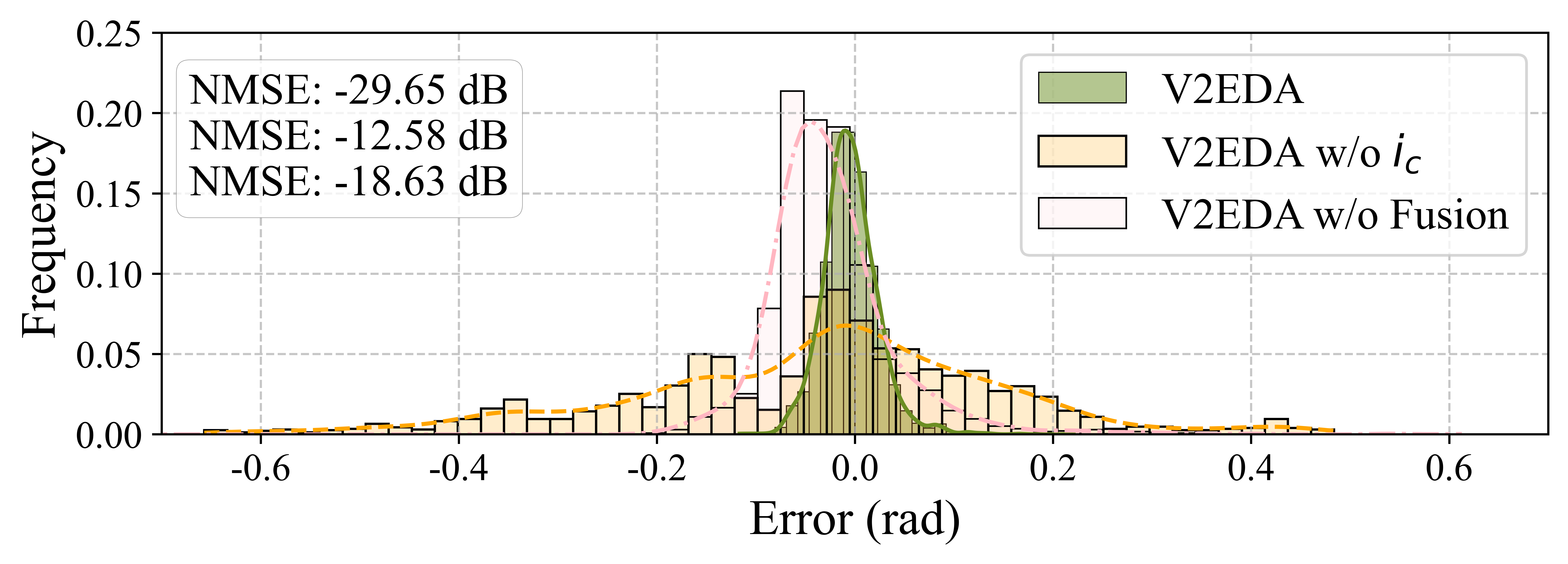}
			\caption{Elevation}
			%\label{fig.7_2}
		\end{subfigure}
		\caption{Angle error in V2EDA.}
		\label{fig.7}
	\end{figure}
	
	We evaluate V2EDA and conduct an ablation to quantify component contributions. Models are trained for 100 epochs with learning rate $10^{-3}$ using Adam. {We compare:
	\begin{enumerate}
	\renewcommand{\labelenumi}{\alph{enumi}.}
	\item V2EDA: full V2EDA.
	\item V2EDA w/o $\mathbf{i}_\text{c}$: V2EDA removes the cropped patch $\mathbf{i}_\text{c}$, retaining only the detector’s bounding-box features, limiting depth cues.
	\item V2EDA w/o  Fusion: V2EDA replaces the Feature Fusion (CA) with a common concatenation fusion strategy.
	\end{enumerate}
	}
	Fig.~\ref{fig.7} visualizes azimuth/elevation estimates (radians) relative to the ISAC BS on the test set, together with error histograms. Azimuthal alignment is stable, with inter-modal discrepancies largely confined to $[-0.02,0.02]$~rad. Elevation is less accurate yet tracks the overall trend. This gap reflects limited monocular depth from below-horizon views: reduced depth observability penalizes elevation more than azimuth. By extracting features from $\mathbf{i}_\text{c}$, the proposed method recovers implicit distance cues, partially closing the elevation deficit. {Moreover, the V2EDA variant with common concat fusion exhibits larger estimation deviation than the original CA-based version, demonstrating that the designed attention structure effectively enhances cross-modal feature alignment and mapping quality.}

	\subsection{MMFE Results}
	\label{sec.6.2}
	
	\begin{figure*}
		\centering
		\includegraphics[width=1\textwidth]{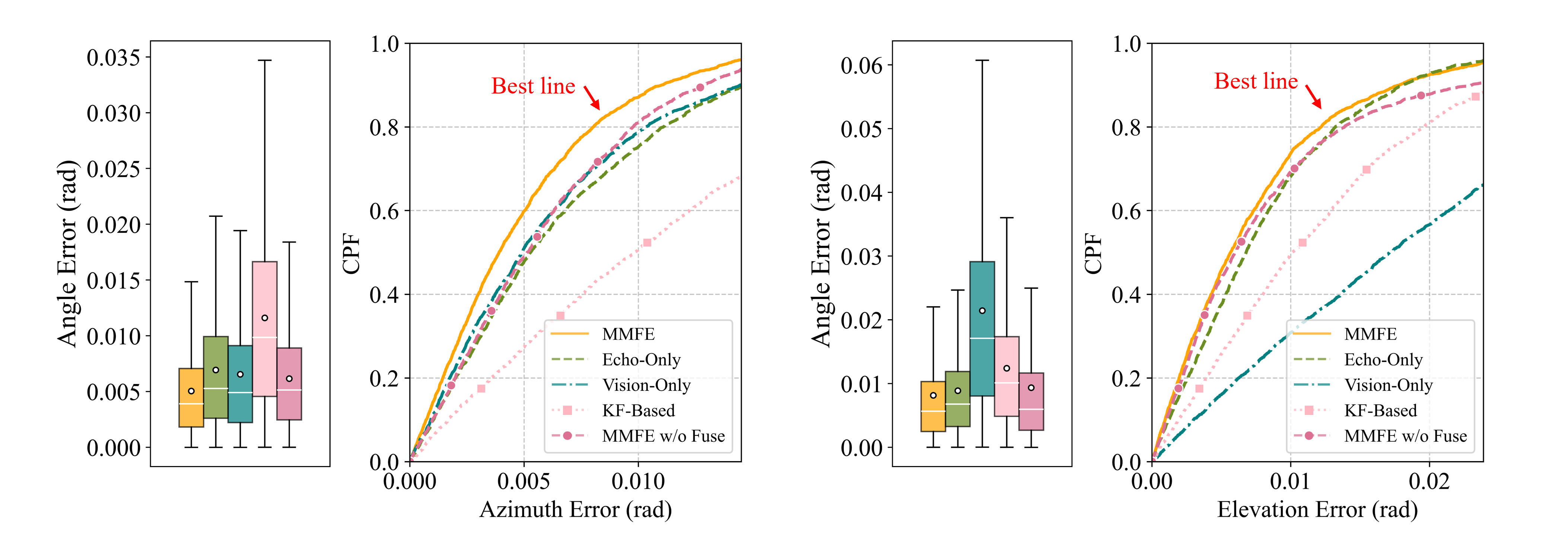}
		\caption{{MMFE angle-error evaluation: The left panel shows the boxplot, and the right panel displays the CPF plot. The two plots share the same legend.}}
		\label{fig.8}
	\end{figure*}

	We analyze MMFE in terms of accuracy, robustness, and complexity. Unless otherwise stated, training uses 100 epochs, learning rate $10^{-3}$, Adam optimizer. To isolate fusion gains we compare:
	{\begin{enumerate}
		\renewcommand{\labelenumi}{\alph{enumi}.}
		\item Vision-Only: V2EDA estimates angles from current VSI.
		\item Echo-Only: MMFE ablated to use historical ESI.
		\item MMFE: Full MMFE using VSI and ESI.
	\item KF-Based: Kalman filter based predictors\cite{4103555, 9145366}, using historical ESI.
		\item MMFE w/o Fuse: MMFE ablated to use simple multimodal fusion rules, using VSI  and ESI.
		%\item MMFE: Full MMFE using VSI and ESI with synchronization errors.
	\end{enumerate}
	}
	
	\textbf{Accuracy Analysis}: 
	Fig.~\ref{fig.8} reports boxplots and cumulative probability functions (CPF) . {In the boxplots on the left, the proposed MMFE exhibits the best performance, with the lowest mean outliers and the most favorable quantile distributions for both azimuth and elevation. In the CPF plots on the right, the ``Best line'' curve also corresponds to MMFE. By contrast, MMFE w/o Fuse performs significantly worse than MMFE and is close to Echo-Only, indicating that the Fuse module effectively combines historical ESI with current VSI to improve angle prediction accuracy, while simple fusion strategies fail to provide comparable benefits.}
	
	{Moreover, the KF-based method exhibits similar error levels for azimuth and elevation because it relies on a model-driven estimation process where both angular states are governed by the same dynamic model and noise assumptions. As a result, the estimation errors tend to be statistically similar across the two dimensions.
	In contrast, learning-based methods show larger discrepancies between azimuth and elevation errors, which is attributed to the data-driven nature of these models, where the learned mapping is influenced by the statistical characteristics of the training dataset.  }
	
	Furthermore, a noteworthy observation is that the Multimodal model achieves a significant reduction in azimuth estimation error compared to the Echo-Only model, whereas the improvement in elevation estimation is relatively limited. This result aligns with the inherent characteristics of each modality: visual perception excels in azimuth estimation but has limited accuracy in elevation. Consequently, the model cannot extract elevation information of the same quality from visual inputs as it does for azimuth. This is not a shortcoming of the model; rather, it demonstrates its intelligence. The fusion model adaptively leverages the complementary strengths of different modalities, relying on a modality when it is reliable and reducing dependence when it is weak, thereby achieving optimal system-level estimation performance.
	
	\begin{figure*}
		\centering
		\includegraphics[width=1\linewidth]{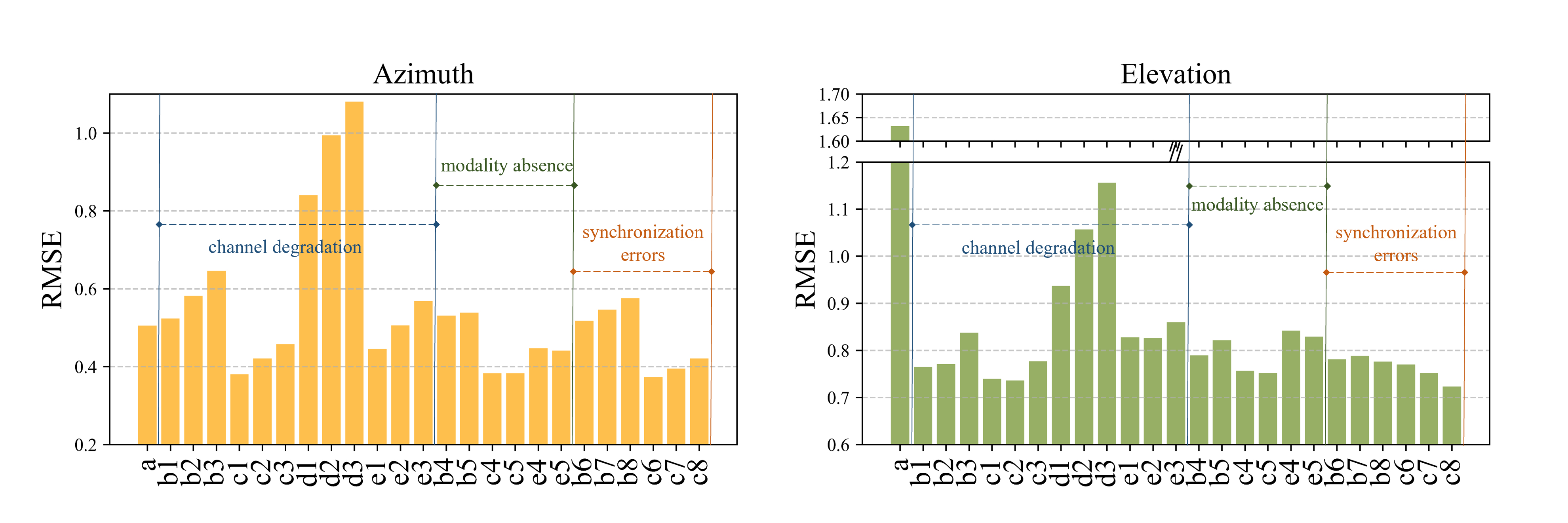}
		\caption{{Fault tolerance capability of MMFE: (a) Vision-Only; (b1)-(b3) Echo-Only at SNR=0,-1,-2; (c1)-(c3) MMFE at SNR=0,-1,-2; (d1)-(d3) KF-Based at SNR=0,-1,-2; (e1)-(e3) MMFE w/o Fuse at SNR=0,-1,-2; (b4)-(b5) Echo-Only with few/many echo losses; (c4)-(c5) MMFE with few/many echo losses; (e4)-(e5) MMFE w/o Fuse with few/many echo losses; (b6)-(b8) Echo-Only time offset with $[-t,0], [-t,t], [0,t]$; (c6)-(c8) MMFE time offset with $[-t,0], [-t,t], [0,t]$. }}
		\label{fig.9}
	\end{figure*}
	
	\textbf{{Robustness Analysis}}: 
	{ Fig.~\ref{fig.9} reports the robustness analysis, including parameter generalization, modality absence, and camera synchronization errors. }
		
	{For parameter generalization, the UAV flight altitude ranges from 30 to 400 m and the velocity ranges from 0 to 27 m/s, providing sufficient diversity in UAV motion states. Channel variations are primarily modeled through different echo signal SNR conditions. Therefore, we evaluate methods using RMSE under channel degradation.} Channel quality is swept by SNR $\in{0,-1,-2}$~dB. The proposed MMFE scheme consistently achieves the best angle estimation performance across all tested channel conditions. As the channel conditions deteriorate (i.e., with decreasing SNR, (b1)-(b3), (c1)-(c3), {(d1)-(d3) and (e1)-(e3)}), the RMSE of all models increases. For example, for the Echo-Only model, the RMSE rises from 0.52 to 0.62 in azimuth and from 0.76 to 0.83 in elevation. {For the KF-Based method, the RMSE rises from 0.84 to 1.08 in azimuth and from 0.9 to 1.15 in elevation.} For the proposed MMFE model, it increases only slightly, from 0.38 to 0.42 in azimuth and from 0.73 to 0.77 in elevation. Owing to the stable information provided by the visual modality, the MMFE model consistently maintains a lower error level.
		
	{For modality absence, the image dataset inherently contains partial visual missing caused by occlusions. Hence, this experiment mainly considers the absence of the echo modality.}	In simulation, sudden data loss is emulated by randomly dropping a small (“few”) or large (“many”) portion of ESI. In Echo-Only, missing ESI is imputed with the last valid sample. In MMFE, imputation is modality-aware: for azimuth, missing ESI is complemented by VSI; for elevation, the most recent valid ESI is preferred. {KF-based operates purely on historical angle estimates and does not rely on multimodal feature inputs. Therefore, modality absence scenarios mainly affect the proposed MMFE framework rather than the KF baseline.} As shown in Fig.~\ref{fig.9}, the proposed MMFE scheme consistently achieves the best angle estimation performance across all sudden failure scenarios.  The robustness advantage of the MMFE model becomes even more pronounced when dealing with varying degrees of sudden data loss ((b4)/(b5), (c4)/(c5),  and (e4)/(e5)). {For example, when the echo data is partially missing, the RMSE of the Echo-Only model increased by 0.01/0.02 in azimuth and 0.03/0.04 in elevation.} This demonstrates that a single modality learning-based is highly vulnerable to the loss of its own data integrity. 
	{While MMFE increased by 0.00/0.00 in azimuth and 0.01/0.01 in elevation. With the direct supplementation of the visual modality, MMFE w/o Fuse method even had a lower RMSE.} The results confirm that the multi modality effectively compensates for degraded echo inputs, allowing the MMFE model to sustain significantly higher robustness. This proves that the MMFE fusion mechanism can effectively compensate for the temporary failure of a single modality, thereby ensuring the persistent reliability of the perception system.

	{For synchronization errors, a temporal offset $\Delta t$ ms is introduced between VSI and ESI acquisition, allowing the echo to occur earlier or later than the visual frame. Three cases are simulated: echo delay $[-t,0]$, random offset $[-t,t]$, and echo advance $[0,t]$, where the maximum offset $t$  corresponds to half of the image frame period.}
	{By comparing (b1) with (b6)–(b8) and (c1) with (c6)–(c8), it can be observed that both the Echo-Only and MMFE methods exhibit good robustness to synchronization errors. This robustness mainly stems from the temporal continuity of the echo measurements and the learning model’s ability to tolerate moderate temporal perturbations in sequential observations. Specifically, The maximum performance degradation of Echo-Only is only 0.05/0.03 in azimuth/elevation RMSE, while the maximum degradation of MMFE is 0.04/0.03.
	Interestingly, the azimuth estimation slightly benefits from echo delay, while the elevation estimation performs better under echo advance. On the one hand, moderate temporal offsets may occasionally improve performance because they implicitly compensate for small synchronization deviations between sensing modalities or introduce a mild temporal smoothing effect in sequential observations. On the other hand, this phenomenon is also related to the different motion dynamics of the two angular dimensions. In UAV scenarios, azimuth variations are typically faster due to dominant horizontal motion, whereas elevation changes are relatively smoother. A slight echo delay therefore provides a temporally smoothed observation that stabilizes azimuth estimation, while a small echo advance yields measurements closer to the current elevation state, which can be beneficial for elevation prediction.}
	
	\begin{table}[htbp] 
		\centering 
		\caption{Model Parameters} 
		\label{tab.1} 
		\begin{tabular}{ccccc}
			\toprule  
			Methods & Parameters  & FLOPs & Model Size & Latency  \\
			\midrule 
			Vision-Only & 9.226 K & 290.328 K & 0.04 MB & 0.91 ms\\
			Echo-Only  & 2.545 K  & 27.456 K & 0.01 MB & 0.51 ms \\
			MMFE & 2.665 K  & 28.080 K  & 0.01 MB & 1.40 ms  \\
			MMFE w/o Fuse & 2.653 K  & 27.984 K  & 0.01 MB & 0.66 ms \\
			\bottomrule 
		\end{tabular} 
	\end{table}
	
	\textbf{Complexity Analysis}: {In addition, we evaluate the computational complexity and inference latency of different methods, as summarized in Table~\ref{tab.1}. The proposed MMFE maintains a lightweight structure, with only marginal increases in parameters (4.72\%) and FLOPs (2.27\%) compared to Echo-Only, and the same model size.  
		
	From a system-level perspective, the overall processing delay consists of vision processing, model inference, and sensing operations. The proposed MMFE introduces an inference latency of 1.40 ms, which is slightly higher than Echo-Only due to multimodal fusion and temporal encoding.
	However, this latency remains a small fraction of the available time budget. For example, under a typical 30 fps vision system (about 33 ms per frame), the MMFE inference accounts for less than 5\% of the frame duration, and therefore does not pose a real-time processing bottleneck.	}

	\subsection{CC-ISAC Results}
	\label{sec.6.3}

	\textbf{Task A:} On the camera side, UAV detection (Task A) is performed using the established YOLOv4\cite{bochkovskiy2020yolov4}, achieving 98\% accuracy on this dataset. The processed visual outputs transmitted to the ISAC BS are mapped according to Eq.~(\ref{eq.9}) and ~(\ref{eq.10}).
	
	%	\begin{figure}
		%	\centering
		%	\includegraphics[width=1\linewidth]{png/fig5_1.png}
		%	\caption{Performance of Vision-assisted Beam Selection}
		%	\label{fig.10}
		%\end{figure}
		
		\begin{figure}[htbp]
			\centering
			% 左子图
			\subfloat[Candidate accuracy]{
				\includegraphics[width=0.47\linewidth]{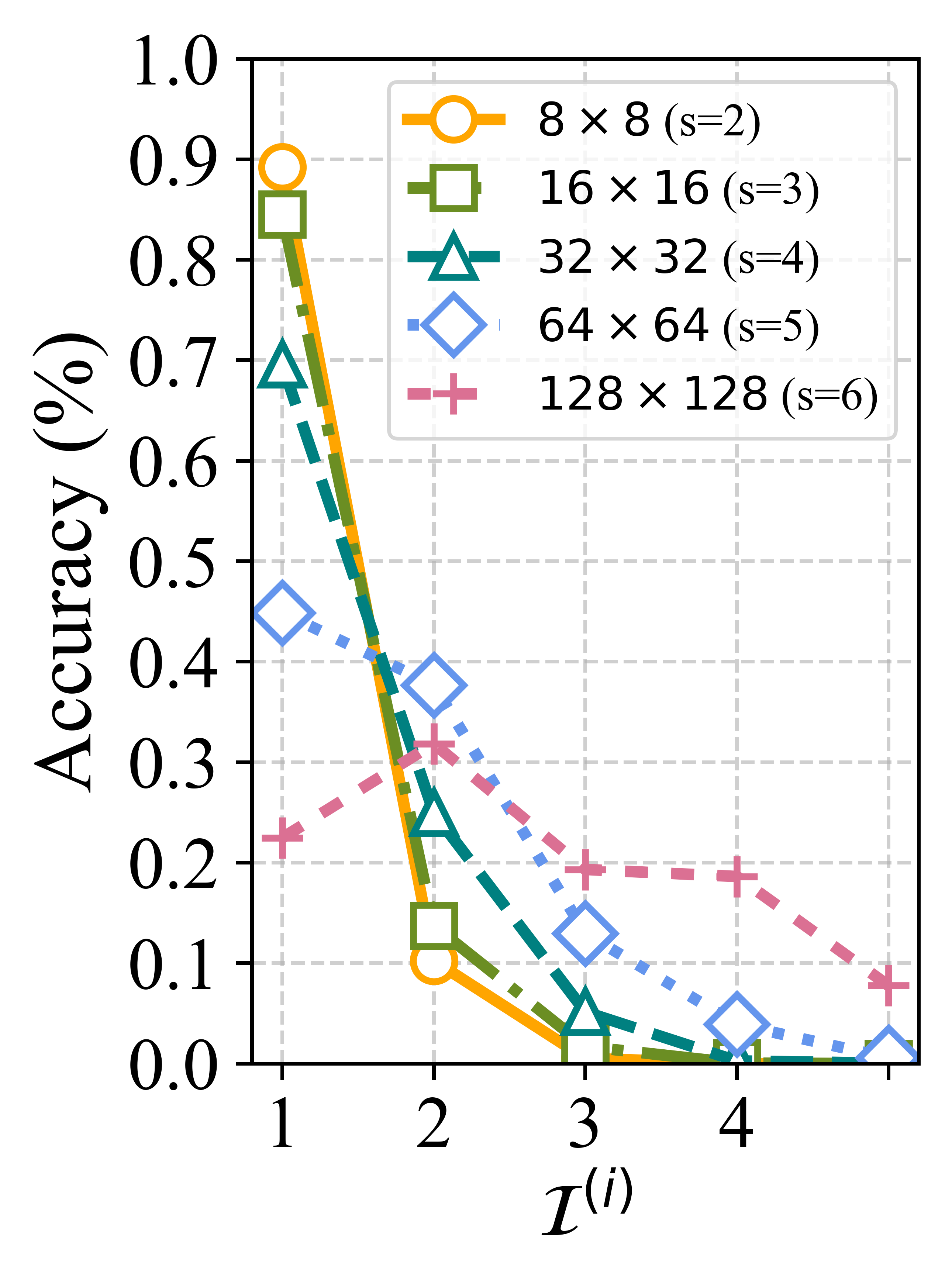}
			}
			\hfill   % 让两图之间留一点空隙
			% 右子图
			\subfloat[Top-k candidate accuracy]{
				\includegraphics[width=0.47\linewidth]{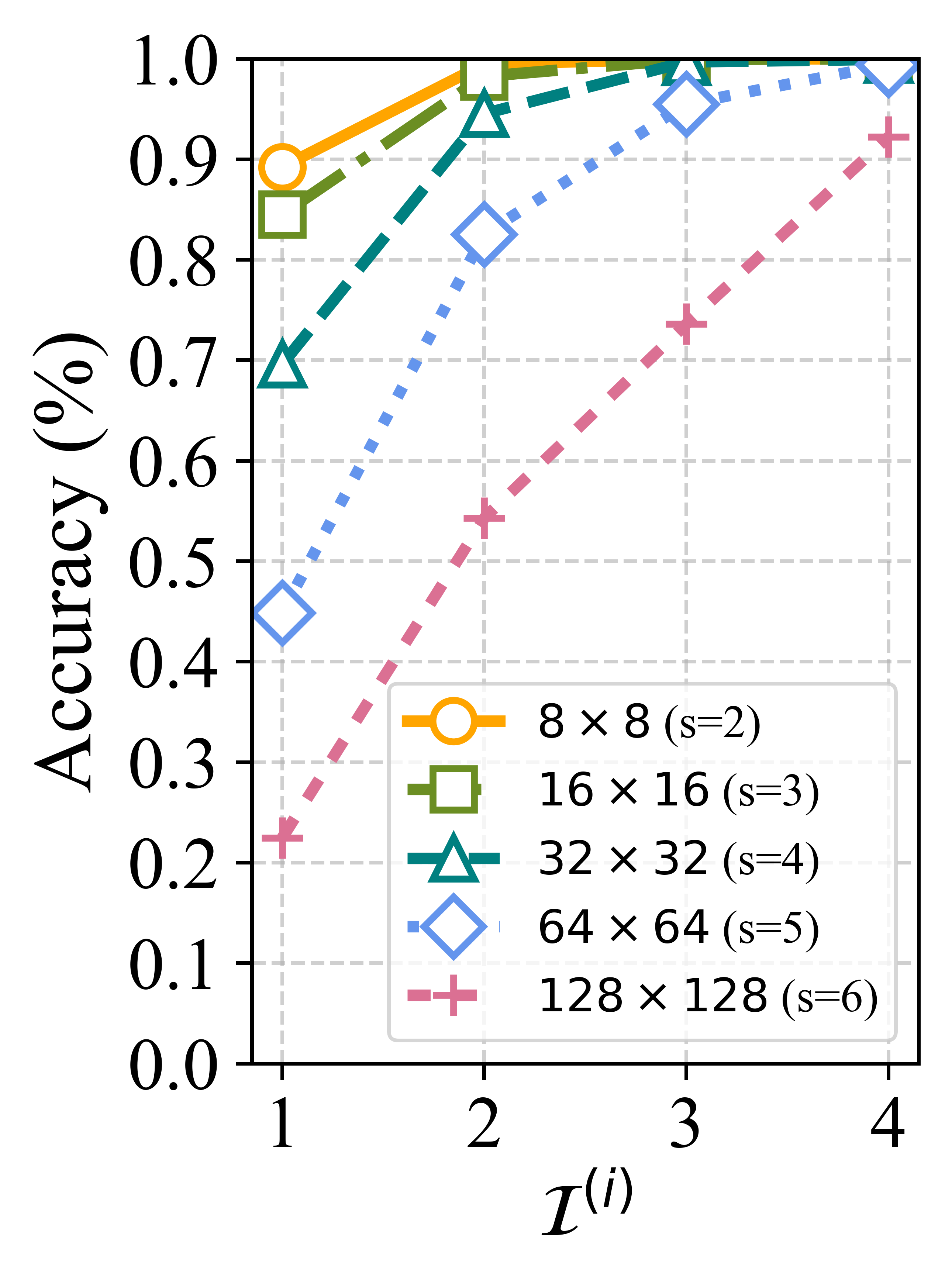}
			}
			
			\caption{Performance of vision-assisted beam selection.}
			\label{fig.10}
		\end{figure}
		\textbf{Task B:} We analyze the performance of the Vision-assisted beam selection method in the predefined codebook. The diffusive beam candidate set is employed, scanned in the order $\mathcal{I}^{(1)} \rightarrow \mathcal{I}^{(2)} \rightarrow \mathcal{I}^{(3)}\rightarrow \mathcal{I}^{(4)}$. As exhibited in Fig.~\ref{fig.10}, we adopted different codebooks with level $s$ (noted their corresponding array number), and showed the accuracy of different candidate sets. Specifically, the x-axis denoted as the candidate set $\mathcal{I}^{(i)}$ calculated using the angles $(\hat{\theta}^{(\text{v})}, \hat{\phi}^{(\text{v})})$  (Eq.~(\ref{eq.13})) estimated by the V2EDA model. and the value on $\mathcal{I}^{(i)}$ indicate the frequency at which the UAV is within the scanning range of $\mathcal{I}^{(i)}$.

		\begin{table*}[htbp] 
			\centering 
			\caption{Beam Steering Efficiency (Beam Scanning Overhead)} 
			\label{tab.2} 
			\begin{tabular*}{0.97\linewidth}{@{\extracolsep{\fill}} cccccc}
				\toprule 
				
				Method & $8 \times 8$ (s=2) & $16 \times 16$ (s=3) & $32 \times 32$ (s=4) & $64 \times 64$ (s=5) & $128 \times 128$ (s=6) \\
				\midrule 
				%	& Exhaustive  search & 5 & 7.5 & 10 & 12.5 & 15 \\
				Single ISAC: Hierarchical scan & 5 & 7.5 & 10 & 12.5 & 15 \\
				
				\midrule
				CC-ISAC: V2EDA  & {1.2908 / -74.18\%} & {1.4634 / -80.49\%} & {2.0048 / -79.95\%} & {3.4936 / -72.05\%} & {7.2173 / -51.51\%} \\
				\bottomrule 
			\end{tabular*} 
		\end{table*}

		%& SNR=0 & {1.1003 / -2.70\%} & {1.1297 / -3.18\%} & {1.3680 / -5.51\%} & {1.8767 / -4.92\%} & {2.9875 / -8.59\%} \\
		%& SNR=-1 & {1.1218 / -3.38\%} & {1.1556 / -3.96\%} & {1.3709 / -5.83\%} & {1.8909 / -4.57\%} & {3.1129 / -8.74\%} \\
		%& SNR=-2 & {1.1312 / -4.86\%} & {1.1628 / -5.76\%} & {1.3739 / -7.30\%} & {1.9077 / -6.95\%} & {3.3198 / -11.15\%} \\
		%& Few lost & {1.1013 / -2.70\%} & {1.1330 / -3.34\%} & {1.3868 / -5.53\%} %& {1.8901 / -4.93\%} & {3.0246 / -8.87\%} \\
		
		%& Lot lost & {1.1144 / -1.69\%} & {1.1460 / -2.56\%} & {1.3914 / -4.63\%} %& {1.9520 / -4.23\%} & {3.1761 / -6.17\%} \\

		It can be observed that the higher the codebook level $s$, the finer the beamwidth, the more concentrated the energy, and the smaller the coverage area of a single scan. 
		In Fig.~\ref{fig.10}(a), the overall trend is that the higher the priority of the candidate set, the higher the accuracy rate of finding UAVs. When $s=2$, the VSI $(\hat{\theta}^{(\text{v})}, \hat{\phi}^{(\text{v})})$ has an 89\% probability of locating the UAV at one time, and the top three candidate sets can find the UAV 100\%. When $s=3$, the VSI $(\hat{\theta}^{(\text{v})}, \hat{\phi}^{(\text{v})})$ has an 84\% probability of locating the UAV at one time, and for the four candidate sets, it can find the UAV 100\%. When $s=4$ and $5$, the width of the beam decreases, and the overall coverage range of top three candidate sets becomes even smaller. There is still a 99.5\% probability of finding the UAV within four candidate sets. When $s=6$, the corresponding beam width is very narrow at this time, and Vision can only provide a rough estimate. Therefore, the trend deviates from the previously observed pattern, and still, 7.6\% of the UAVs cannot be found within four candidate sets. 
		
		For a clearer observation of the overall trend, Fig.~\ref{fig.10}(b) summarizes the cumulative accuracy of the top-$k$ candidate sets. Specifically, the $i$-th bar denotes the probability that the UAV can be successfully located within the first $i$ candidate sets. The results are consistent with Fig.~\ref{fig.10}(a), while offering a more intuitive view of the proposed method’s beam selection performance.
		
		Overall, although there are errors in visual Angle prediction, it can limit beam selection to a small subset in most cases, thereby transforming exhaustive search into an efficient local search. This indicates that the Vision-assisted beam selection method can help BS find UAVs quickly. However, the result of this coarse positioning is more suitable for searching with wider beams.
		% 我们提供了采用Vision-assisted beam alignment方法估计的UAV角度在预定义码本上的表现情况。如eq13-16表明的那样，我们的码本索引扫描顺序按照集合I1-I2-I3-I4。如8展示了测试集中不同码本等级s（对应标明了常见的天线数量阵列）下，I1-I2-I3-I4和粗扫描失败的频率。其中I_i轴对应的值表示估计的角度（a,b）计算出来的码本索引集合I_i的扫描范围能够覆盖到真实UAV的频率。
		% 首先，我们容易知道，码本等级s越高，意味着波束越细，能量越集中，单次扫描覆盖的范围越小。可以看到，整体趋势是，优先级越高的索引集合找到UAV的准确率越高，在s=2时，视觉感知结果(Sv)有89%的概率能一次就定位到UAV，I1-I2-I3集合的能100%找到UAV；在s=3时，视觉感知结果(Sv)有84%的概率能一次就定位到UAV，I1-I2-I3-I4集合的能100%找到UAV；在s=4和5时，波束的宽度变小，I1I2I3的总体覆盖范围更小了，此时依旧有99.5%的概率能在I1-I2-I3-I4内找到UAV；当s=6时，此时对应的波束宽度很窄，而Vision只能提供一个粗估计，因此结果并符合之前的逐步下降的规律，并且仍有7.6%的UAV不能在I1-I2-I3-I4内找到。
		% 可以看到，尽管视觉角度预测存在误差，但它在大多数情况下能将波束选择限制在一个小的子集内，从而将 exhaustive 搜索转化为一个高效的局部搜索。这表明，Vision-assisted beam alignment方法能够帮助BS快速的找到UAV，不过这种粗定位的结果更适合较宽的波束来搜寻。
		
		% 为量化波束对准效率，波束扫描开销是根据波束搜索过程中消耗的平均扫描次数来评估的。该过程基于初始角度估计启动，在成功探测目标时视为完成，从而直接反映了初始访问期间花费的时间和能量资源。。如表2所示，随着波束宽度的变窄，搜索开销增大，与传统分层扫描方法相比，本文提出的视觉辅助搜索策略将平均扫描次数急剧减少了71%。这一显著的降低意味着，在搜索阶段，所提方法仅需不到三分之一的扫描操作即可实现目标锁定，极大地提升了初始接入效率。
		
		To quantify the beam steering efficiency, the beam scanning overhead is evaluated in terms of the average number of scans consumed in the beam search process by Eq.~(\ref{eq.over}). This process is initiated based on an initial angle estimate and is considered complete upon successful target detection, thereby directly reflecting the time and energy resources expended during initial access. As shown in Table~\ref{tab.2}, the scanning overhead increases as the beam-width narrows. Compared to the conventional hierarchical scanning method, our proposed vision-assisted search strategy sharply reduces the average number of scans by 71\%. This remarkable reduction implies that the proposed method requires less than one-third of the scanning operations to accomplish target acquisition during the search phase, thereby significantly enhancing the initial access efficiency.

		\begin{table*}[htbp] 
			\centering 
			\begin{threeparttable}
				
				\caption{Beam Tracking Efficiency (Beam Scanning Overhead)} 
				
				\label{tab.3} 
				\begin{tabular*}{0.98\linewidth}{@{\extracolsep{\fill}} ccccccc}
					\toprule 
					
					Method &Case& $8 \times 8$ (s=2) & $16 \times 16$ (s=3) & $32 \times 32$ (s=4) & $64 \times 64$ (s=5) & $128 \times 128$ (s=6)  \\
					\midrule 
					\multirow{5}{*}{Single ISAC: Echo-Only} 
					%	& Exhaustive  search & 5 & 7.5 & 10 & 12.5 & 15 \\
					& SNR=0 & {1.1308 } & {1.1668 } & {1.4478 } & {1.9738} & 3.2682 \\
					& SNR=-1 & {1.1611 } & {1.2032 } & {1.4557 } & {1.9814} & 3.4110 \\
					& SNR=-2 & {1.1890 } & {1.2339 } & {1.4821 } & {2.0501} & 3.7365 \\
					& Few lost & {1.1319 } & {1.1722 } & {1.4679 } & {1.9882} & 3.3192 \\
					& Lot lost & {1.1336 } & {1.1761 } & {1.4590 } & {2.0383} & 3.3848 \\
					\midrule
					\multirow{3}{*}{Single ISAC: KF-Based} 
					%	& Exhaustive  search & 5 & 7.5 & 10 & 12.5 & 15 \\
					& SNR=0 & {1.2418 } & {1.3325 } & {1.6288 } & {2.3809} & 4.7797 \\
					& SNR=-1 & {1.3266 } & {1.4267 } & {1.7496 } & {2.6232} & 5.6807 \\
					& SNR=-2 & {1.3682 } & {1.4904 } & {1.8447 } & {2.8739} & 6.1463 \\
					%& Few lost & {1.2298 } & {1.3231 } & {1.6166 } & {2.3552} & 4.6999 \\
					%& Lot lost & {1.2385 } & {1.3172 } & {1.6144 } & {2.3170} & 4.5815 \\

					\midrule
					\multirow{5}{*}{CC-ISAC: MMFE*}  
					& SNR=0 & {1.1003 / -2.70\%} & {1.1297 / -3.18\%} & {1.3680 / -5.51\%} & {1.8767 / -4.92\%} & {2.9875 / -8.59\%} \\
					& SNR=-1 & {1.1218 / -3.38\%} & {1.1556 / -3.96\%} & {1.3709 / -5.83\%} & {1.8909 / -4.57\%} & {3.1129 / -8.74\%} \\
					& SNR=-2 & {1.1312 / -4.86\%} & {1.1628 / -5.76\%} & {1.3739 / -7.30\%} & {1.9077 / -6.95\%} & {3.3198 / -11.15\%} \\
					& Few lost & {1.1013 / -2.70\%} & {1.1330 / -3.34\%} & {1.3868 / -5.53\%} & {1.8901 / -4.93\%} & {3.0246 / -8.87\%} \\
					& Lot lost & {1.1144 / -1.69\%} & {1.1460 / -2.56\%} & {1.3914 / -4.63\%} & {1.9520 / -4.23\%} & {3.1761 / -6.17\%} \\
					
					%& SNR=0 & {1.1003  } & {1.1297 } & {1.3680 } & {1.8767 } & {2.9875 } \\
					%	& SNR=-1 & {1.1218 } & {1.1556 } & {1.3709 } & {1.8909 } & {3.1129 } \\
					%	& SNR=-2 & {1.1312 } & {1.1628 } & {1.3739  } & {1.9077  } & {3.3198 } \\
					%	& Few lost & {1.1013 } & {1.1330 } & {1.3868  } & {1.8901 } & {3.0246 } \\
					
					%	& Lot lost & {1.1144} & {1.1460} & {1.3914 } & {1.9520} & {3.1761 } \\
					\bottomrule 
				\end{tabular*} 
				\begin{tablenotes}[flushleft]  % flushleft 左对齐，默认居中
					\footnotesize  % 字号缩小，符合IEEE规范
					\item[*] The percentages following the slash denote the overhead reduction rate compared to the Single ISAC: Echo-Only baseline, calculated as $(\text{Overhead}_{\text{MMFE}}-\text{Overhead}_{\text{Echo-Only}})	/ \text{Overhead}_{\text{Echo-Only}} \times 100 \%$.
				\end{tablenotes}
			\end{threeparttable}
		\end{table*}
		\begin{figure*}
			\centering
			\includegraphics[width=1\linewidth]{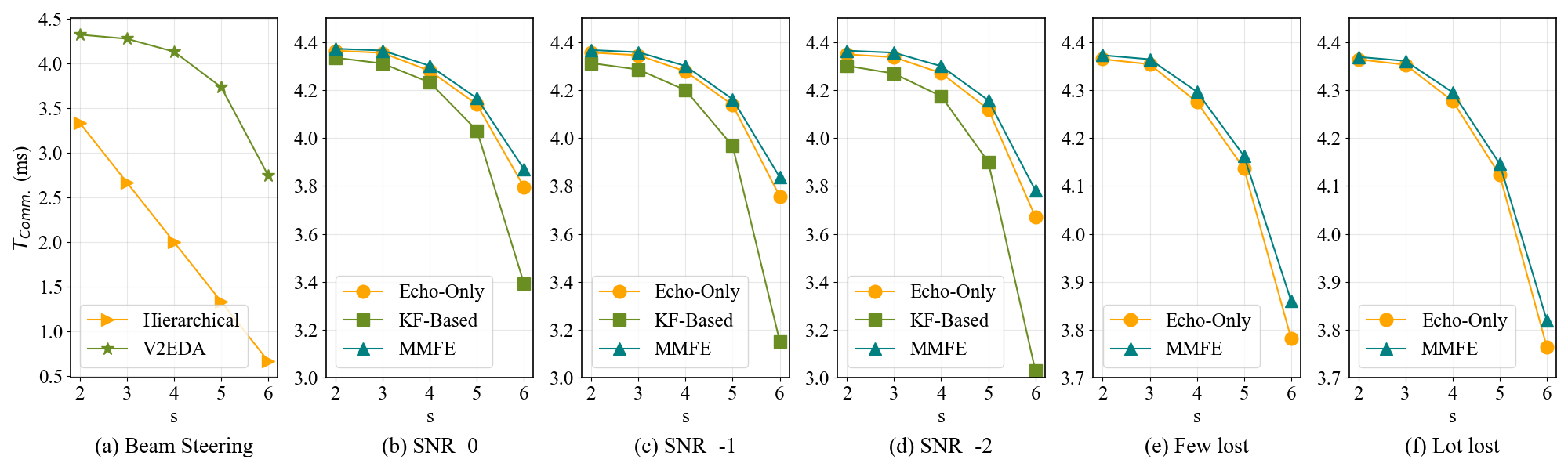}
			\caption{{$T_{\text{Comm.}}$ resource recovery in half-frame: (a) Beam Steering; (b)-(f) Beam Tracking.}}
			\label{fig.11}
		\end{figure*}

		% 本文提出框架
		\textbf{Task C:} % 我们分析在波束追踪阶段，采用MMFE算法进行角度估计对于系统的增益，类似的，我们分析了the beam scanning overhead作为感知增益的体现,并且采用与图9相同的情况模拟。结果如表3所示，CC-ISAC在不同的天线数量（码本精度要求）和不同的case下，开销都更少，开销降低在-1.69~11.15\%不等。
		To evaluate the system gain of the proposed MMFE algorithm during the beam tracking phase, we employ the beam scanning overhead as a key performance metric, which directly reflects the improvement in communication resource efficiency achieved through enhanced sensing accuracy. Simulations were conducted under the same channel conditions and sudden failure scenarios as in Fig.~\ref{fig.9}, with the results summarized in Table~\ref{tab.3}.
		
		The data demonstrate that the proposed CC-ISAC scheme consistently achieves lower beam scanning overhead across varying numbers of antennas (i.e., different codebook resolution requirements) and diverse test cases. {Specifically, Compared with Echo-Only, the MMFE achieved overhead reduction ranges from 1.69\% to 11.15\%. Compared with KF-Based, MMFE achieved overhead reduction ranges from 11.39\% to 45.99\%}.
		
		%\textbf{Consistent Performance Improvement:} 
		The universally positive results demonstrate that the MMFE algorithm reliably enhances beam tracking efficiency. Fusing multimodal data, it provides more accurate and robust angle estimates, which in turn narrows the beam search space and accelerates target re-acquisition.
		
		%\textbf{Context-Dependent Efficacy:} 
		The variation in the degree of improvement reflects the adaptive nature of the fusion mechanism. The highest gains are typically realized in challenging scenarios where single-modality perception is degraded, allowing the complementary strengths of the fused modalities to shine.  The smaller, yet still positive, gains in other cases confirm that the framework introduces no performance regression while adding crucial robustness.
		
	{	\textbf{Communication Resource Gain:} 
		To quantify the practical impact of sensing overhead on communication efficiency, we map the beam scanning process to a realistic 5G NR frame structure.
		Specifically, each radio frame has a duration of 10 ms, consisting of 10 subframes. Each subframe contains one slot with 14 OFDM symbols. In NR beam management, a single synchronization signal block beam occupies 4 consecutive OFDM symbols. 
		In practical deployments, beam sweeping is typically required to be completed within a half-frame duration (i.e., 5 ms). Therefore, the number of OFDM symbols consumed by beam scanning directly determines the remaining time-domain resources available for communication. Fig. \ref{fig.11} illustrates the resulting communication time $T_{\text{Comm.}}$ under different beam management strategies for both beam steering and beam tracking, where $T_{\text{Comm.}}$ is defined in Eq.~(\ref{eq.22}).
		
		As shown in Fig. Fig. \ref{fig.11}(a), the proposed V2EDA-based beam steering achieves a substantial increase in available communication time compared to the conventional hierarchical search.
		This gain primarily stems from the significant reduction in beam search space. In the hierarchical strategy, beam sweeping follows a multi-layer search process, where a large number of wide-to-narrow beams must be sequentially tested. This results in a high sensing overhead that consumes a considerable portion of the half-frame duration.
		In contrast, the V2EDA model provides a coarse but reliable prior of the UAV direction, enabling the beam search to be confined to a small set of candidate beams. As a result, the number of required beam scans is drastically reduced, leading to a substantial decrease in sensing time and a corresponding increase in communication resource availability.
		
		For beam tracking, Fig. \ref{fig.11}(b)–(f), the room for further reducing sensing time is inherently limited compared to beam steering, and thus the improvement in communication time appears relatively modest. Nevertheless, such reductions can translate into noticeable gains in long-term system throughput, especially in continuous operation scenarios.
		Specifically, tracking dominates in persistent sensing scenarios. During beam tracking, the beam is already aligned in the previous time step, and only incremental adjustments are required.  
		The proposed MMFE method improves tracking accuracy by leveraging multimodal information and demonstrates consistent performance across different channel conditions and failure scenarios. This reduces the probability of beam misalignment and avoids occasional re-scanning.
		
		Overall, the results confirm that the proposed framework is particularly effective in reducing the initial access overhead, while still providing consistent gains during the tracking phase, thereby improving the overall sensing–communication efficiency of the ISAC system.
		}

		\section{Conclusion}
		\label{sec.7}
		%In this paper, we have presented the Camera-Cooperative ISAC (CC-ISAC) framework, a novel multimodal fusion approach that effectively addresses the critical challenge of non-cooperative UAV detection in resource-constrained ISAC environments.   Through the strategic coordination of visual perception and ISAC sensing, our framework establishes a complementary paradigm where cameras provide coarse-grained airspace monitoring while ISAC delivers precise, targeted sensing capabilities.
		
		%The technical contributions of this work are threefold.   First, the proposed Vision-to-Echo Data Alignment (V2EDA) model successfully bridges the geometric disparity between heterogeneous sensors, enabling accurate mapping from visual coordinates to beamspace.   Second, the Multimodal Fusion-based Estimation (MMFE) model effectively integrates historical multimodal data with current observations through gated fusion and temporal modeling, achieving robust motion state estimation.   Third, our comprehensive evaluation on the DeepSense6G dataset demonstrates substantial performance gains, including a 71\% reduction in beam selection overhead and 1.69-11.15\% improvement in tracking efficiency, while maintaining high angular accuracy.
		
		{In this paper, we proposed a Camera-Cooperative ISAC (CC-ISAC) framework for efficient beam steering and tracking of non-cooperative UAVs in resource-constrained ISAC systems. By reallocating sensing tasks across heterogeneous modalities, the framework employs vision for coarse airspace awareness and ISAC for fine-grained sensing, thereby improving sensing efficiency while reducing ISAC resource overhead.}
		
		{To enable practical multimodal cooperation, we developed the Vision-to-Echo Data Alignment (V2EDA) model for visual-to-beamspace mapping and the Multimodal Fusion-based Estimation (MMFE) model for robust motion state estimation through cross-modal fusion. Extensive experiments on the DeepSense6G dataset demonstrate that the proposed framework achieves high angular accuracy while reducing beam selection overhead by 71\% and improving beam tracking efficiency by 1.69--11.15\%.}
		
		%The CC-ISAC framework fundamentally transforms the traditional trade-off between S\&C functions in ISAC systems.   By intelligently allocating sensing resources based on multimodal information, our approach not only ensures reliable UAV detection but also liberates significant resources for communication tasks.   This work thus represents a significant step toward practical ISAC deployment for low-altitude security applications, demonstrating that multimodal fusion can effectively break the resource competition bottleneck that has long constrained conventional ISAC designs.
		
		{Future work will focus on hardware-based validation using synchronized camera--radar platforms, where practical factors such as clutter, calibration errors, and non-line-of-sight (NLoS) propagation can be systematically investigated.} We will also extend the framework to more complex scenarios, including dense UAV environments and advanced evasion behaviors, while further improving the computational efficiency for real-time deployment.  The integration of additional modalities and the development of adaptive fusion strategies present promising directions for enhancing the robustness and scalability of multimodal ISAC systems.

		\bibliographystyle{IEEEtran}
		\bibliography{Bibliography}

	\end{document}